\documentclass[10pt]{fairmeta}
\usepackage{fontspec}
\usepackage{graphicx}
\usepackage{caption}
\usepackage[numbers,sort&compress]{natbib}
\usepackage{amsmath,amssymb,booktabs,tabularx,array}
\usepackage{sectsty}
\usepackage{pdflscape,colortbl}
\usepackage[hidelinks]{hyperref}
\usetikzlibrary{arrows.meta}
\newfontfamily\reportfont[AutoFakeBold=1.5]{Outfit-Regular.ttf}
\newcommand{\reportheading}[1]{{\color{metablue}\reportfont #1}}
\allsectionsfont{\reportheading}
\newcolumntype{L}{>{\raggedright\arraybackslash}X}
\title{{\reportfont\color{black}\fontsize{14}{18}\selectfont
  HoloMotion-1.5: Training a 0.4B MoE Transformer for Humanoid\\
  Motion Tracking on 10,000+ Hours of Data within 100 GPU-Hours}}
\author[]{Yucheng Wang}
\author[]{Min Du}
\author[]{Liu Liu}
\author[\dagger]{Zhizhong Su}
\affiliation[]{Horizon Robotics}
\code{\textcolor{metablue}{\url{https://github.com/HorizonRobotics/HoloMotion}}}
\webpage{\textcolor{metablue}{\href{https://horizonrobotics.github.io/robot_lab/holomotion_1_5/}{\nolinkurl{https://horizonrobotics.github.io/robot_lab/holomotion_1_5}}}}
\correspondence{Zhizhong Su at \textcolor{metablue}{\email{zhizhong.su@horizon.auto}}}

\abstract{HoloMotion-1 demonstrated zero-shot humanoid tracking with a 0.4B-parameter sparse MoE Transformer trained on over 2,000 hours of motion. Training required about 9,200 GPU-hours~\citep{holomotion1}, limiting adaptation to new applications. To make large-scale tracker training more accessible, HoloMotion-1.5 introduces a three-stage framework that separates shared control, residual specialization, and global routing. Groups learn residual specialists in parallel on the same frozen generalist and evaluate only their own experts during imagined rollouts~\citep{supermimic}. We fix expert weights and fit a global router and gate that read only the reference. Compact streaming supports roughly five times as much motion data, over 10,000 hours. Training from prepared references is budgeted at 83 active GPU-hours, less than 1\% of the previous GPU time. The tracker sets new bests on all three benchmark tracks. It achieves 90.36\% success on held-out HumanTracker, highest among compared policies without global pose inputs. Full-motion XY-aligned errors on \mbox{MotionChallenge} are 25.88~mm on LAFAN1 and 17.66~mm on BONES-SEED, 53\% and 36\% below prior published bests.}

\begin{document}
\newgeometry{hmargin=1in,top=0.74in,bottom=0.70in}
\maketitle
\enlargethispage{15pt}

\noindent\begin{minipage}{\linewidth}
\centering
\begingroup
\definecolor{diagramink}{HTML}{173047}
\definecolor{diagramgray}{HTML}{506576}
\definecolor{diagramline}{HTML}{CBDAE5}
\definecolor{diagramblue}{HTML}{1854A8}
\definecolor{diagramexpert}{HTML}{117466}
\definecolor{diagramgate}{HTML}{96521F}
\definecolor{diagrambluefill}{HTML}{F0F5FC}
\definecolor{diagramexpertfill}{HTML}{EAF6F2}
\definecolor{diagramgatefill}{HTML}{FCF1E8}
\definecolor{diagramneutral}{HTML}{F7F9FC}
\begin{tikzpicture}[
    x=1cm, y=1cm, >=latex,
    every node/.style={text=diagramink, font=\reportfont\fontsize{7.9}{9.5}\selectfont},
    panel title/.style={anchor=west, inner sep=0pt, font=\reportfont\bfseries\fontsize{9.5}{11}\selectfont},
    panel detail/.style={anchor=east, text=diagramgray, inner sep=0pt},
    panel mark/.style={fill=diagramink, text=white, rounded corners=1pt, inner sep=0pt,
      minimum width=0.33cm, minimum height=0.33cm, font=\reportfont\bfseries\footnotesize},
    stage frame/.style={draw=black!18, rounded corners=4pt, line width=0.5pt},
    stage name/.style={anchor=west, inner sep=0pt, text=black!90, font=\reportfont\bfseries\fontsize{8.3}{10}\selectfont},
    training op/.style={rounded corners=2pt, minimum width=4.35cm, minimum height=0.53cm,
      inner ysep=2pt, align=center},
    expert unit/.style={rounded corners=2pt, minimum width=0.84cm, minimum height=0.44cm,
      inner sep=2pt, align=center},
    footer note/.style={text=diagramgray, font=\reportfont\fontsize{7.4}{9}\selectfont},
    model block/.style={draw=diagramline, rounded corners=3pt, fill=white, align=center,
      inner ysep=3pt, minimum height=0.58cm, minimum width=2.55cm, line width=0.45pt},
    data path/.style={->, draw=diagramgray, line width=0.55pt},
    transfer path/.style={->, draw=diagramblue, line width=0.70pt},
    scale path/.style={->, draw=diagramgate, line width=0.70pt},
    short arrow/.style={-{Latex[length=2.0pt,width=1.8pt]}},
    stage bus/.style={draw=diagramgray, line width=0.65pt},
    stage branch/.style={stage bus, -{Latex[length=2.6pt,width=2.5pt]}},
    arithmetic/.style={circle, draw=diagramline, fill=white, minimum size=0.40cm, inner sep=0pt}
]
    \path[use as bounding box] (0,0.18) rectangle (16.34,-12.10);

    \begin{scope}[yshift=1.62cm]
    \node[panel mark] at (0.18,-1.62) {a};
    \node[panel title] at (0.48,-1.62) {Stagewise training};
    \draw[stage frame] (0,-1.97) rectangle (5.06,-5.31);
    \draw[stage frame] (5.64,-1.97) rectangle (10.70,-5.31);
    \draw[stage frame] (11.28,-1.97) rectangle (16.34,-5.31);

    \node[stage name] at (0.24,-2.29) {1\enspace Shared generalist learning};
    \node[training op,fill=diagrambluefill] (refs) at (2.53,-2.90) {Stream compact motion pages};
    \node[training op,fill=diagrambluefill] (sim) at (2.53,-3.63) {Robot simulation and replay};
    \node[training op,fill=diagrambluefill,text=diagramblue] (base) at (2.53,-4.36)
      {\textbf{Train one shared generalist}};
    \draw[data path,short arrow] (refs.south) -- (sim.north);
    \draw[transfer path,short arrow] (sim.south) -- (base.north);
    \node[footer note] at (2.53,-4.99) {Reused frozen in Stages 2 and 3};

    \node[stage name] at (5.88,-2.29) {2\enspace Residual specialist learning};
    \node[training op,fill=diagramexpertfill] (partition) at (8.17,-2.85) {Partition motions into \(N\) groups};
    \node[expert unit,fill=diagramexpertfill] (groupone) at (6.64,-3.70) {\(1\)};
    \node[expert unit,fill=diagramexpertfill] (grouptwo) at (7.66,-3.70) {\(2\)};
    \node[expert unit] at (8.68,-3.70) {\(\cdots\)};
    \node[expert unit,fill=diagramexpertfill] (grouplast) at (9.70,-3.70) {\(N\)};
    \node[training op,fill=diagramexpertfill,text=diagramexpert] (bank) at (8.17,-4.60)
      {\textbf{Learn residuals in parallel}};
    \coordinate (split bus) at (8.17,-3.25);
    \coordinate (merge bus) at (8.17,-4.12);
    \draw[stage bus] (partition.south) -- (split bus);
    \draw[stage bus] (groupone.north |- split bus) -- (grouplast.north |- split bus);
    \foreach \groupnode in {groupone,grouptwo,grouplast}{
      \draw[stage branch] (\groupnode.north |- split bus) -- (\groupnode.north);
      \draw[stage bus,draw=diagramexpert] (\groupnode.south) -- (\groupnode.south |- merge bus);
    }
    \draw[stage bus,draw=diagramexpert] (groupone.south |- merge bus) -- (grouplast.south |- merge bus);
    \draw[stage branch,draw=diagramexpert] (merge bus) -- (bank.north);
    \node[footer note] at (8.17,-5.09) {\(E\) experts + a local router per group};

    \node[stage name] at (11.52,-2.29) {3\enspace Global routing and gating};
    \node[training op,fill=diagramgatefill] (teachers) at (13.81,-2.90) {Stack experts and distill routers};
    \node[training op,fill=diagramgatefill,text=diagramgate] (global) at (13.81,-3.63)
      {\textbf{Anneal global routing to Top-\(K\)}};
    \node[training op,fill=diagramgatefill] (schedule) at (13.81,-4.36)
      {Freeze router and fit residual gate};
    \draw[data path,short arrow] (teachers.south) -- (global.north);
    \draw[data path,short arrow] (global.south) -- (schedule.north);
    \node[footer note] at (13.81,-4.99) {All \(N\cdot E\) specialists stay frozen};

    \draw[transfer path] (5.08,-3.63) -- (5.60,-3.63);
    \draw[scale path] (10.72,-3.63) -- (11.24,-3.63);

    \node[panel mark] at (0.18,-5.75) {b};
    \node[panel title] at (0.48,-5.75) {Sparse Transformer at inference};
    \node[panel detail] at (16.32,-5.75) {Shared generalist expert + \(K\) residual specialists};

    \node[model block,fill=diagramneutral,minimum width=2.12cm] (reference) at (1.47,-6.66) {Reference\\motion};
    \node[model block,fill=diagramexpertfill] (features) at (4.50,-6.66)
      {Reference feature \(r\)};
    \node[model block,fill=none,shade,left color=diagramexpertfill,right color=diagramgatefill] (router) at (8.11,-6.66)
      {Reference-only router\\Top-\(K\) indices and weights};
    \node[model block,fill=diagramgatefill] (gate) at (12.40,-6.66)
      {Residual gate\\\(g(r)\)};
    \node[model block,fill=diagramneutral,minimum width=2.12cm] (proprioception) at (1.47,-8.44)
      {Robot\\proprioception};
    \node[model block,fill=diagrambluefill] (transformer) at (4.60,-8.27)
      {Causal self-attention\\Observation history};
    \node[model block,fill=diagramexpertfill,minimum width=3.15cm] (residual experts) at (8.11,-7.82)
      {Residual specialist experts\\Weighted mixture};
    \node[model block,fill=diagrambluefill,minimum width=3.15cm] (shared) at (8.11,-8.90)
      {Shared generalist expert};
    \node[arithmetic] (multiply) at (12.40,-7.82) {\(\times\)};
    \node[arithmetic] (sum) at (13.76,-8.27) {\(+\)};
    \node[model block,fill=none,shade,top color=diagramgatefill,bottom color=diagrambluefill,minimum width=0cm,text width=1.70cm] (head) at (15.30,-8.27)
      {Action head\\Joint targets};

    \begin{scope}[
      >={Latex[length=2.45pt,width=2.1pt]},
      inference line/.style={draw=black,line width=0.60pt},
      inference arrow/.style={inference line,->}
    ]
    \coordinate (reference split) at (2.82,-6.66);
    \draw[inference line] (reference.east) -- (reference split);
    \draw[inference arrow] (reference split) -- (features.west);
    \draw[inference arrow] (features.east) -- (router.west);
    \draw[inference arrow] (features.north) -- (4.50,-6.04) -| (gate.north);
    \draw[inference arrow] (router.south) -- (residual experts.north);
    \draw[inference arrow] (reference split) |- ([yshift=0.23cm]transformer.west);
    \fill[black] (reference split) circle[radius=0.75pt];
    \draw[inference arrow] (proprioception.east) -- ([yshift=-0.17cm]transformer.west);
    \draw[inference line] (transformer.east) -- (6.16,-8.27);
    \fill[black] (6.16,-8.27) circle[radius=0.75pt];
    \draw[inference arrow] (6.16,-8.27) |- (residual experts.west);
    \draw[inference arrow] (6.16,-8.27) |- (shared.west);
    \draw[inference arrow] (residual experts.east) -- (multiply.west);
    \draw[inference arrow] (gate.south) -- (multiply.north);
    \draw[inference arrow] (multiply.east) -| (sum.north);
    \draw[inference arrow] (shared.east) -| (sum.south);
    \draw[inference arrow] (sum.east) -- (head.west);
    \end{scope}
    \end{scope}

    \begin{scope}[yshift=1.00cm]
    \node[panel mark] at (0.18,-8.96) {c};
\node[panel title] at (0.48,-8.96) {Tracking benchmarks};
\node[panel detail] at (16.32,-8.96) {HoloMotion-1.5 and published methods};

\tikzset{
  benchmark name/.style={anchor=west,inner sep=0pt,font=\reportfont\bfseries\fontsize{8.15}{9.5}\selectfont},
  benchmark metric/.style={anchor=east,inner sep=0pt,text=diagramgray,font=\reportfont\fontsize{7.2}{8.5}\selectfont},
  benchmark tick/.style={anchor=east,inner sep=0pt,text=diagramgray,font=\reportfont\fontsize{6.3}{7.2}\selectfont},
  benchmark value/.style={anchor=south,inner sep=0.7pt,font=\reportfont\fontsize{6.0}{7.0}\selectfont},
  benchmark model/.style={anchor=east,rotate=60,inner sep=0pt,font=\reportfont\fontsize{6.5}{7.6}\selectfont}
}
\newcommand{\benchmarkpanel}[6]{%
  \def\benchmarkbase{-2.32}%
  \def\benchmarkheight{1.38}%
  \draw[stage frame] (0,-0.22) rectangle (5.22,-4.02);
  \node[benchmark name] at (0.20,-0.49) {#1};
  \node[benchmark metric] at (5.02,-0.49) {#2};
  \draw[diagramline,line width=0.4pt] (0.53,\benchmarkbase) -- (5.10,\benchmarkbase);
  \foreach \tick in {#3,#4,#5}{%
    \pgfmathsetmacro{\tickheight}{\benchmarkbase+\benchmarkheight*(\tick-#3)/(#5-#3)}
    \draw[diagramline,line width=0.35pt] (0.50,\tickheight) -- (0.57,\tickheight);
    \node[benchmark tick] at (0.46,\tickheight) {\tick};
  }%
  \ifnum#3>0
    \draw[diagramgray,line width=0.45pt] (0.53,-2.36) -- (0.59,-2.27)
      (0.60,-2.36) -- (0.66,-2.27);
  \fi
  \foreach \score/\model [count=\barindex from 0] in {#6}{%
    \pgfmathsetmacro{\barcenter}{0.86+0.573*\barindex}
    \pgfmathsetmacro{\barheight}{\benchmarkbase+\benchmarkheight*(\score-#3)/(#5-#3)}
    \pgfmathsetmacro{\valuelift}{0.05*mod(\barindex,2)}
    \ifnum\barindex=0
      \def\barcolor{diagramblue}\def\labelcolor{diagramblue}
    \else
      \def\barcolor{diagramline}\def\labelcolor{diagramgray}
    \fi
    \fill[\barcolor] ({\barcenter-0.187},\benchmarkbase) rectangle ({\barcenter+0.187},\barheight);
    \node[benchmark value,text=\labelcolor] at (\barcenter,{\barheight+\valuelift}) {\score};
    \node[benchmark model,text=\labelcolor] at (\barcenter,{\benchmarkbase-0.14}) {\model};
  }%
}
\begin{scope}[shift={(0,-9.00)}]
  \benchmarkpanel{HumanTracker}{Success (\%) \(\uparrow\)}{80}{86}{92}{
    90.36/\textbf{HoloMotion-1.5},89.56/SONIC 1.1,89.40/SONIC,88.16/HEFT,
    86.52/ScaleBFM-M,86.52/ScaleBFM-XL,85.16/GRIT,82.24/HoloMotion 1.4.0}
\end{scope}
\begin{scope}[shift={(5.56,-9.00)}]
  \benchmarkpanel{LAFAN1}{Full-motion XY (mm) \(\downarrow\)}{0}{60}{120}{
    25.88/\textbf{HoloMotion-1.5},54.86/ScaleBFM-M,64.17/GRIT 99k,65.16/MimicLite,
    69.25/HEFT PMG,90.78/SONIC Orig.,118.61/SONIC LL,119.10/SONIC 1.1}
\end{scope}
\begin{scope}[shift={(11.12,-9.00)}]
  \benchmarkpanel{BONES-SEED}{Full-motion XY (mm) \(\downarrow\)}{0}{20}{40}{
    17.66/\textbf{HoloMotion-1.5},27.41/HEFT PMG,30.62/MimicLite,30.81/ScaleBFM-M,
    32.64/GRIT 99k,33.48/SONIC Orig.,34.12/SONIC LL,35.40/HoloMotion 1.3.2}
\end{scope}

    \end{scope}
\end{tikzpicture}
\endgroup


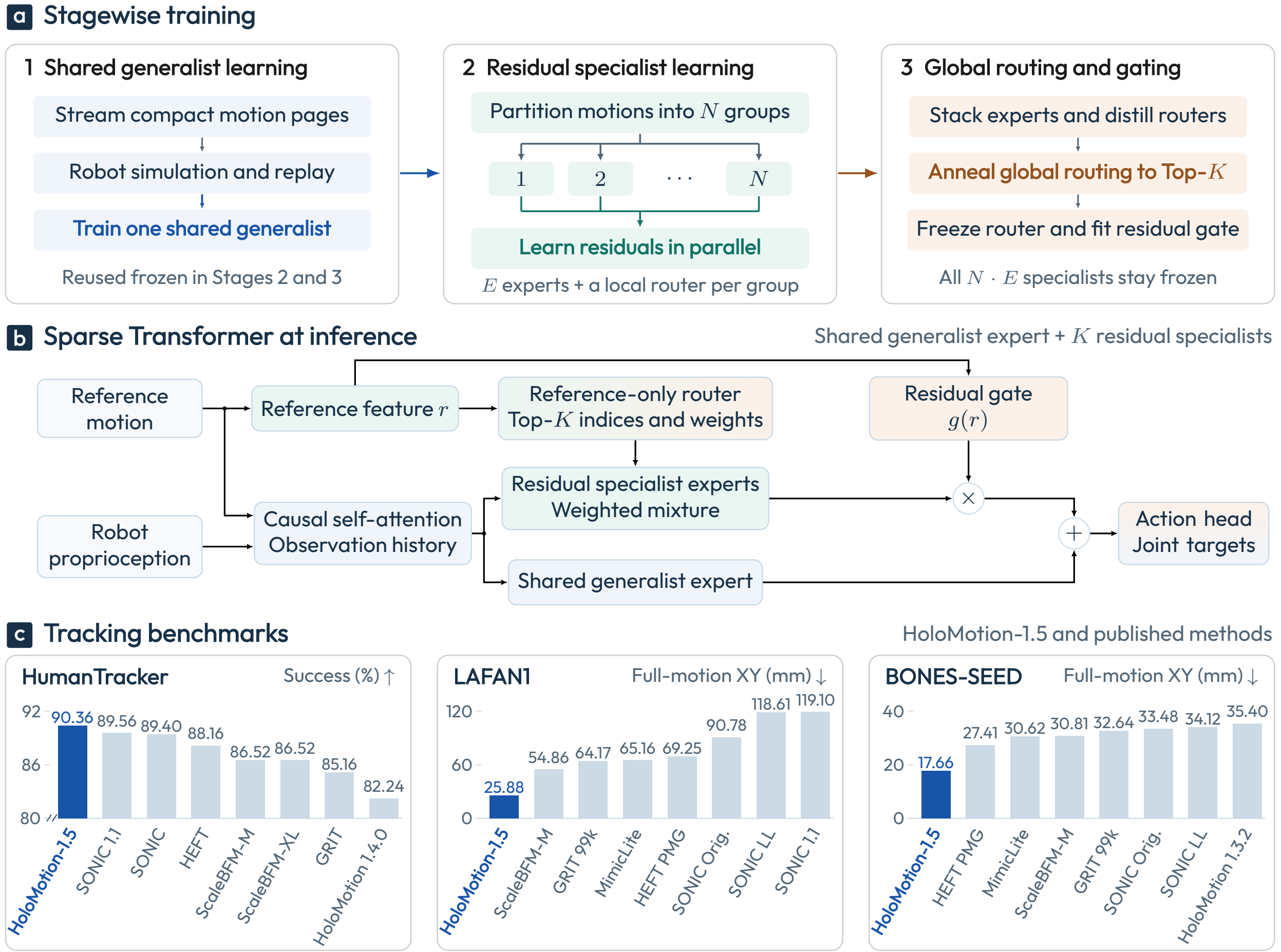
\captionof{figure}{\textbf{(a)} Train one shared generalist, learn residual specialists in parallel groups on the frozen generalist, then fit the global router and gate with the expert bank frozen. \textbf{(b)} The reference selects two specialists and sets the gain on their combined correction. \textbf{(c)} Highest HumanTracker success without global pose inputs and lowest full-motion XY-aligned error on both MotionChallenge tracks.}
\label{fig:recipe}
\end{minipage}
\restoregeometry

\section{Introduction}

HoloMotion-1 showed that a sparse MoE Transformer could track unseen whole-body motions on a real robot after training on studio capture, teleoperation, and reconstructed video~\citep{holomotion1}. Training that controller required substantial GPU time. Motion catalogs now span more styles, contacts, and transitions, giving a single controller more opportunities to learn skills it can reuse across motions. The next challenge is to expand the repertoire and the capacity to learn it while reducing the GPU work required for training and keeping computation predictable at every control step.

Different sources contribute different kinds of motion. Studio capture can cover deliberate transitions with precise measurements. Teleoperation captures what an operator asks a robot to do, and video can add movements that are expensive to record in a studio. A motion collection assembled this way will contain many variations of the same underlying skill. The training recipe should be able to learn what those variations have in common and still leave room for corrections that matter to a particular movement.

Consider a robot following a reference that goes from walking straight to making a sharp turn. Both phases need balance and weight transfer, so they can share a controller for those skills. The turn may also need a different foot placement and a coordinated change in trunk motion. The router sees the turn in the reference and selects specialists that can help. The gate sets the strength of their combined correction and can leave it at zero on motions that the generalist handles well. The actor still uses robot feedback throughout, for example to adjust its joint targets if the actual landing differs from the reference.

Inspired by Heald et al.'s account of contextual inference in motor learning~\citep{heald2021}, we separate learning a skill from deciding when to use it. In their reaching experiments, briefly reintroducing a familiar force brought back the corresponding compensation. Their model explained this recovery by stronger expression of a motor memory that had already been learned. In our controller, the reference supplies the context. It tells the router which corrections to recruit and the gate how strongly to apply them while the generalist keeps supplying feedback.

We organize training into three stages (Figure~\ref{fig:recipe}). In Stage~1, a shared generalist learns whole-body feedback across several motion sources. Every later stage reuses that controller with its weights frozen, so shared control is learned once. In Stage~2, we partition motions by source and shard. Each group learns residual specialists and a local router against the same frozen generalist. The tracking objective lets the specialists divide the work within their group without expert-specific screening or clustering. In Stage~3, we assemble the specialists, distill the local routers into a reference-only global router, and refine that router with rollouts. We then freeze the router and fit the residual gate.

During Stage~2, even dense routing evaluates only the 32 specialists in a group. Localizing these evaluations saves work at every imagined actor step~\citep{supermimic}, and running the groups in parallel shortens elapsed time. Stage~3 can evaluate the assembled bank, including during an initial dense phase, but it skips parameter updates for the generalist and all specialists. Annealed Top-\(K\) Routing gives more experts training signal early and limits deployment to two specialist evaluations per control step. Section~\ref{sec:compute} reports GPU-hours summed over devices separately from elapsed time.

A large catalog also changes how training data has to reach the learner. GPU retargeting converts source motions to robot references. Compact pages store root and joint trajectories and reconstruct link kinematics as needed, so more motions fit in the working set. A background loader prepares new pages while simulation and learning are already running. The simulator and replay keep access to the pages they still need, and released pages make room for new motions. This lets the policy start from the first ready references and continue drawing from a catalog larger than the working memory. Batched dynamics gradients, activation checkpointing, and CUDA Graphs further reduce execution overhead~\citep{xmimic,instanttrack}.

We evaluate transfer and tracking accuracy separately. On held-out HumanTracker, the tracker completes 90.36\% of motions, highest among compared policies without global pose inputs. On MotionChallenge's LAFAN1 and BONES-SEED tracks, whose references are in the training catalog, full-motion XY-aligned errors are 25.88 and 17.66~mm. These are 53\% and 36\% below the previous published bests in our comparison.

\newgeometry{hmargin=1in,top=0.86in,bottom=0.74in}
\setlength{\parskip}{0.42em}
\section{Background}
\label{sec:data}

The motion catalog shapes what a tracker can learn. Its scale also changes how much computation training requires and how references reach the simulator.

\subsection{The rapid growth of motion data}

Web video and purpose-built capture are expanding the range of motions available for training, with different trade-offs in scale, precision, and contact coverage.

\paragraph{Internet video.}
Shared archives such as AMASS and LAFAN1 made existing capture reusable~\citep{amass,lafan1}. Automated person tracking and 3D reconstruction now extract motion from existing recordings, without arranging a capture session for every activity. MotionMillion combines video reconstructions and existing archives for over 2,000 hours of human motion. AnyMo reports 5,048 hours for OmniHuMo, comprising about 3.2 million motion sequences reconstructed from internet videos. Movens-5M adds robot retargeting and execution checks to produce over 10,000 hours of robot references~\citep{motionmillion,omnihumo,movenszero}.

\paragraph{Purpose-built capture.}
Purpose-built sessions can repeat precise contacts across actors and record the object motions that a robot has to reproduce. ChingMu's MotionDecode reports more than 1,000 hours of optical capture with synchronized object poses across households, manufacturing, logistics, and other settings. Noitom Robotics' HiPHI releases 617.5 hours including mirrors, from 308.7 hours of original optical recordings, with synchronized object trajectories for its interaction sequences~\citep{chingmudata,hiphi}. SONIC trained its tracker on references retargeted from 700 hours of in-house human motion~\citep{sonic}. A planned session can record multiple performers and takes of a specific action or measure an object throughout an interaction. Web video adds situations, styles, and transitions that are harder to commission at comparable scale.

\paragraph{Coverage and usable references.}
A larger mixed catalog can expose a tracker to unfamiliar combinations, such as turning into a deep crouch while reaching. HoloMotion-1 demonstrated transfer to dances reconstructed from video on a physical G1 after training on a 2,000+ hour mixed catalog~\citep{holomotion1}. Greater variety also brings uneven coverage, occlusions, and contacts that may be hard to infer from video. Training at this scale requires reliable robot references and a sampler that reaches varied motions within a limited training budget.

ScaleBFM holds simulator collection fixed and compares nested reference sets~\citep{scalebfm}. Tripling the BONES-SEED subset yields marginal gains. Adding complementary datasets improves cross-source tracking. Movens-Zero holds the generator size and the proportions of scenes, motion types, and modalities fixed~\citep{movenszero}. Training with more source videos improves conditional generation and the robot's ability to track generated motions. These results motivate a catalog with complementary sources. For a tracker, the catalog defines which references are available, and rollout sampling determines which ones it sees. Section~\ref{sec:retargeting} describes our source mixture and how we convert it to robot references.

\subsection{MoE model design for efficiency}

Sparse expert activation separates the total capacity of a model from how much computation each input uses. Switch Transformers train with sparse token dispatch, and DeepSeekMoE reserves always-active experts for computations shared across inputs~\citep{switchtransformer,deepseekmoe}. HoloMotion-1 uses sparse routing to bring high model capacity to a controller that must run at each control step~\citep{holomotion1}. An inference-time activation limit leaves open how the experts are learned. Hard routing from the outset directs tracking gradients to the initially selected paths. Dense training lets more experts learn from each sample, with cost that grows with the number of experts considered jointly.

Several lines of work reuse learning across stages. Sparse Upcycling initializes a sparse mixture from a dense checkpoint. Branch-Train-MiX starts from a common model, adapts its branches separately, then fine-tunes a joint router. FlexOlmo trains experts against a common frozen model and assembles their routing embeddings, while Branch-Train-Stitch learns small integration layers around frozen experts~\citep{upcycling,btx,flexolmo,bts}.

\begin{samepage}
In humanoid control, BumbleBee clusters related motions, trains expert policies for the clusters, and distills their behavior into a generalist~\citep{bumblebee}. We retain the specialists in the deployed controller. Each group trains a small mixture of corrective experts on top of the same generalist. Source and shard boundaries divide the workload, and the tracking objective learns the division of labor within each group without additional expert-specific screening or clustering.\par
\end{samepage}

Combining those mixtures raises two questions at deployment. The first is which experts to run. Entmax can put exact zeros on low-scoring experts, with the support determined by the logits. Hard Top-\(K\) sets an explicit limit on active expert computation~\citep{entmax,switchtransformer}. Top-\(K\) sparsemax imposes a support cap directly on the sparsemax projection~\citep{correia2020sparsemarginalization}. The second is how much the chosen experts should change the action. Residual reinforcement learning uses a learned correction on top of an existing controller~\citep{residualrl}. When a new motion has weak matches across the specialist bank, the shared controller may already be the better guide. This motivates learning both a sparse choice and a separate residual strength, using the reference motion to make each decision. Section~\ref{sec:method} describes how the three stages train these components.

\subsection{Data loading and training throughput}

Data throughput matters because rollouts and replay need reference poses even when the full catalog cannot fit on the accelerator. tf.data composes parallel reads, cached transformations, and prefetching. FFCV couples an efficient storage format with asynchronous transfer, and SPDL overlaps array decoding, preprocessing, and GPU transfer~\citep{tfdata,ffcv,spdl}. Robotics pipelines add persistent structure to that data. LeRobotDataset v3 indexes synchronized observations and actions by episode and supports streaming, while Reverb manages sampling from experiences generated by a running agent~\citep{lerobotv3,reverb}.

\begin{samepage}
Reference motions have to stay readable beyond the batch that first loads them. A simulator follows consecutive targets in a clip and may look ahead. After the episode ends, a learner can request those same targets again through replay. Keeping the entire bank resident makes these lookups simple, but memory use and initialization time grow with the bank. A streaming trainer needs to keep clips readable for active episodes and replay while bringing in new motions under a fixed memory budget.
\end{samepage}

\begin{samepage}
SuperMimic improves sample efficiency by differentiating through a learned dynamics model. Its published controller uses MLPs~\citep{supermimic}. Transformer rollouts still incur attention and, under dense routing, many expert evaluations. InstantTrack and XMimic address GPU execution and scheduling, and BlinkMimic directs practice toward failure states~\citep{instanttrack,xmimic,blinkmimic}. BlinkMimic 2 reports fast fitting on BONES-SEED, while \(\nabla\)Mimic studies gradients through the simulator itself~\citep{blinkmimic2,nablamimic}. We adapt imagined actor updates from SuperMimic, batched dynamics gradients and activation checkpointing from XMimic, and CUDA Graph capture from InstantTrack. We further reduce the work itself by learning shared control once, training small expert groups, and freezing the expert bank during global routing.
\end{samepage}

\noindent\begin{minipage}{\linewidth}
\centering
\resizebox{0.94\linewidth}{!}{\begin{tikzpicture}[
  x=1cm,y=1cm,font=\reportfont\small,
  group/.style={rounded corners=3pt,draw=panelstroke,minimum height=0.82cm,
    align=center,inner sep=3pt,font=\reportfont\bfseries\small},
  subhead/.style={font=\reportfont\bfseries\footnotesize,align=center},
  item/.style={anchor=west,font=\reportfont\small},
  cell/.style={anchor=center,minimum width=2.72cm,minimum height=0.48cm,
    text height=0.16cm,text depth=0.05cm,rounded corners=2pt,
    font=\reportfont\footnotesize,inner sep=2pt,align=center},
  learn/.style={cell,fill=teallight,text=tealdark},
  fixed/.style={cell,fill=bluelight,text=bluedark},
  absent/.style={cell,fill=neutrallight,text=black!70},
  detail/.style={font=\reportfont\footnotesize,align=center,text=black!82}
]
\definecolor{panelstroke}{HTML}{CAD5E3}
\definecolor{bluelight}{HTML}{EAF1FB}
\definecolor{bluedark}{HTML}{284B78}
\definecolor{teallight}{HTML}{E3F3EB}
\definecolor{tealdark}{HTML}{185942}
\definecolor{warmlight}{HTML}{F8EBDF}
\definecolor{warmdark}{HTML}{8C4E25}
\definecolor{neutrallight}{HTML}{F3F5F7}

\draw[panelstroke,rounded corners=4pt] (0,2.91) rectangle (15.65,9.82);
\node[item,font=\reportfont\bfseries\small] at (0.34,9.34) {Rollout-based training};
\node[learn,minimum width=1.59cm] at (10.60,9.34) {Trainable};
\node[fixed,minimum width=1.59cm] at (12.35,9.34) {Frozen};
\node[absent,minimum width=1.59cm] at (14.10,9.34) {Absent};

\node[group,fill=bluelight,text=bluedark,minimum width=2.72cm] at (4.65,8.55)
  {Stage 1\\Shared generalist};
\node[group,fill=teallight,text=tealdark,minimum width=2.72cm] at (7.57,8.55)
  {Stage 2\\Residual specialists};
\node[group,fill=warmlight,text=warmdark,minimum width=5.64cm] at (11.95,8.55)
  {Stage 3\\Global routing and residual gating};

\node[subhead] at (4.65,7.92) {One shared actor};
\node[subhead] at (7.57,7.92) {One router per group};
\node[subhead] at (10.49,7.92) {Fit global router};
\node[subhead] at (13.41,7.92) {Fit gate};
\draw[panelstroke] (0.34,7.55) -- (14.88,7.55);

\foreach \y/\name/\sI/\sII/\sIII/\sIV in {
  7.11/{Shared generalist}/learn/fixed/fixed/fixed,
  6.52/{Residual specialists}/absent/learn/fixed/fixed,
  5.93/{Router}/absent/learn/learn/fixed,
  5.34/{Learned gate}/absent/absent/absent/learn,
  4.75/{Critic}/learn/learn/learn/learn,
  4.16/{Dynamics model}/learn/learn/learn/learn}{
    \node[item] at (0.34,\y) {\name};
    \foreach \x/\state in {4.65/\sI,7.57/\sII,10.49/\sIII,13.41/\sIV}{
      \ifnum\pdfstrcmp{\state}{learn}=0
        \node[learn] at (\x,\y) {Trainable};
      \else
        \ifnum\pdfstrcmp{\state}{fixed}=0
          \node[fixed] at (\x,\y) {Frozen};
        \else
          \node[absent] at (\x,\y) {Absent};
        \fi
      \fi
    }
}
\node[item] at (0.34,3.37) {Residual gain};
\node[detail,fill=neutrallight,rounded corners=2pt,minimum width=2.72cm,minimum height=0.64cm] at (4.65,3.37) {No residual};
\node[detail,fill=teallight,rounded corners=2pt,minimum width=2.72cm,minimum height=0.64cm] at (7.57,3.37) {Scheduled\\\(g_{\rm train}:0\to1\)};
\node[detail,fill=teallight,rounded corners=2pt,minimum width=2.72cm,minimum height=0.64cm] at (10.49,3.37) {Scheduled\\\(g_{\rm train}:0\to1\)};
\node[detail,fill=warmlight,rounded corners=2pt,minimum width=2.72cm,minimum height=0.64cm] at (13.41,3.37) {Learned\\\(g(r)\in[0,1]\)};

\end{tikzpicture}
}
\captionof{figure}{Trainable and frozen modules during rollout-based training. The shared generalist is frozen after Stage~1. Residual specialists learn in Stage~2 and stay frozen as the global router and gate are fitted in Stage~3.}
\label{fig:shared-specialist-training}
\end{minipage}

\section{Efficient Training of a Sparse MoE Transformer}
\label{sec:method}

To train a large expert bank without repeating shared control updates or evaluating every expert in each rollout, we separate shared control, groupwise residual specialization, and global routing. All three stages use the same reward, termination rule, and observation layout for rollout-based training. Figure~\ref{fig:shared-specialist-training} shows which parts of the controller learn as the tracking task stays fixed.

\paragraph{Shared reward.}
The reward must balance motion fidelity with smooth joint commands and joint limits. We track poses and velocities, emphasize the head, wrists, and feet, and penalize action changes and soft joint-limit violations. For each tracking term, $E_i$ is the indicated mean squared error, or the squared error for a single target. The total reward at a 20~ms control step is
\begin{equation}
r=0.02\left[\sum_i w_i\exp(-E_i/\sigma_i^2)+\mathbf{1}_{d\leq1}
  -0.1\lVert a_t-a_{t-1}\rVert_2^2-10\sum_j\bigl([q_j-q_j^{+}]_+ + [q_j^{-}-q_j]_+\bigr)\right],
\label{eq:tracking-reward}
\end{equation}
where $q_j^-,q_j^+$ are soft joint bounds and $d$ is the normalized failure severity. For root velocities, $E_i/\sigma_i^2$ denotes the squared local error divided by $(\lVert v_{\mathrm{ref}}\rVert+0.1)^2$, using the corresponding reference linear or angular velocity.

\begin{center}
\small
\begin{tabularx}{\linewidth}{Lrr}
\toprule
Tracking term & Weight & Scale \\
\midrule
Pelvis-frame link position & 1 & 0.30 m \\
Pelvis-frame link orientation & 1 & 0.40 rad \\
World-frame link linear velocity & 1 & 1.00 m/s \\
World-frame link angular velocity & 1 & 3.14 rad/s \\
Pelvis-frame root linear velocity & 2 & Reference-dependent \\
Pelvis-frame root angular velocity & 2 & Reference-dependent \\
Pelvis-frame head proxy, wrists, and feet & 5 & 0.10 m \\
World-frame pelvis XY & 1 & 0.50 m \\
World-frame pelvis height & 1 & 0.30 m \\
World-frame link orientation & 1 & 0.40 rad \\
\bottomrule
\end{tabularx}
\end{center}

The link terms cover the pelvis, torso, and paired hip-roll, knee, ankle, shoulder-roll, elbow, and wrist-yaw links. The first four terms use ankle pitch. The world-orientation and extremity terms use ankle roll. The head proxy is 0.5~m along the torso's up direction. Orientation errors use squared relative rotation angles.

\paragraph{Termination.}
We use torso tilt to detect tracking failure. Let $d=|R_{\mathrm{torso},zz}-R^{\mathrm{ref}}_{\mathrm{torso},zz}|/0.8$. Physical rollouts terminate on tracking failure when $d>1$. Clip completion and the episode step limit are timeouts. In imagined rollouts we use $\operatorname{sigmoid}((d-1)/0.05)$ as the failure probability and weight future rewards by cumulative survival.

\paragraph{Actor observations.}
The actor needs feedback to correct its state and lookahead to anticipate the next movement. Each of its eight history tokens has 383 values. We adapt the features from HEFT~\citep{heft} and use SuperMimic's five-frame reference stride~\citep{supermimic}, sampling offsets \(1,6,11,16\) from the token frame. Proprioception contributes 93 values (29 joint positions relative to the nominal pose, 29 joint velocities, body-frame angular velocity, projected gravity, and the previous 29-dimensional action). The current reference-to-robot root rotation contributes six. Each of four future poses contributes 71 values (reference-relative translation, six-dimensional robot-relative rotation, reference joints, joint-angle error, reference root height, and reference projected gravity). Future indices are clamped to the end of the current clip. Actor inputs contain no measured root translation or linear velocity.

\paragraph{Critic and router observations.}
A two-token Transformer critic also sees root height and linear velocity, robot state, and reference-relative joint, link, and root errors. Its reference token includes future root displacement, orientation, linear and angular velocity, and joint positions and velocities. Where routing or gating is trained, both use only reference-relative displacement, reference joint positions, height, and projected gravity at the four future offsets.

\subsection{Shared generalist learning}
We learn reusable feedback skills on six motion libraries. Figure~\ref{fig:stage2-motion-gallery} illustrates these sources alongside OmniHuMo, which enters specialist training.

\paragraph{Actor and critic updates.}
To learn more from each batch of simulated experience, we adapt SuperMimic's imagined actor updates to the shared Transformer~\citep{supermimic}. We fit one-step dynamics over eight recorded actions, then differentiate through 24 imagined policy steps with reused history keys and values (Section~\ref{sec:execution}). The critic fits \(\lambda\)-returns and bootstraps the final imagined state. Deployment uses the actor alone.

The generalist trains on one GPU for 16 active hours, with a 250K-step cap. LAFAN1 receives twice the reset weight of each of the other five libraries, giving them probabilities of \(2/7\) and \(1/7\), respectively. We use the 150K vectorized-step checkpoint as the starting point for specialization.

\subsection{Residual specialist learning}
We pool smaller libraries and divide larger ones into disjoint groups. Within a large library, we assign whole shards in decreasing frame count to the group with the fewest frames so far. Each of the \(N\) groups trains \(E\) specialists and a reference-only router on its own manifest, reusing the same frozen Stage~1 generalist and normalization statistics. The feed-forward block adds \(g_{\mathrm{train}}\sum_i w_i f_i(x)\) to the generalist output. The scheduled gain \(g_{\mathrm{train}}\) scales the combined correction. The sum covers only the group's \(E\) experts even during dense routing. Each group fits a fresh critic and dynamics model to transitions collected on its own motion partition.

\begin{samepage}
\paragraph{Annealed Top-\(K\) Routing.}
To give all experts training signal early and bound deployment cost, we anneal routing from dense softmax through an entmax blend to hard Top-\(K\). Let \(p\) be the elapsed fraction of the active training budget. Routing progress is \(s=u((p-p_0)/(p_1-p_0))\), with \(u(v)=3\bar v^2-2\bar v^3\) and \(\bar v=\operatorname{clip}_{[0,1]}(v)\). At \(p_0\) the dense phase ends, and at \(p_1\) hard routing begins. For logits \(z\), let \(q_K(z)\) be the softmax over the \(K\) highest logits, with zeros elsewhere. We blend it with \(\alpha\)-entmax~\citep{entmax}.
\begin{equation}
w(z,s)=(1-s)\,\operatorname{entmax}_{1+s}(z)+s\,q_K(z).
\label{eq:annealed-routing}
\end{equation}
\end{samepage}

For \(s>0\), entmax assigns weight \([1+s(z_i-\tau_s)]_+^{1/s}\) to expert \(i\), where \([v]_+=\max(v,0)\) and \(\tau_s\) normalizes the weights. As \(s\to0^+\), \((1+sv)^{1/s}\to e^v\) and \(\tau_s\to\log\sum_j e^{z_j}\), giving softmax. The weights sum to one throughout the blend. The entmax term can support more than \(K\) experts for \(s<1\). At \(s=1\), only \(q_K\) remains, with exactly \(K\) nonzero weights (Figure~\ref{fig:annealed-routing}). In Stage~2, the residual gain \(g_{\mathrm{train}}=u(p)\) rises from zero to one. In Stage~3, the router fit ramps up the correction, then a separate fit learns the gain from reference features.

We partition the seven-library catalog into \(N=32\) groups with \(E=32\) experts each (Figure~\ref{fig:stage2-motion-gallery}). Table~\ref{tab:corpus} reports per-source rollout coverage. Each group has a two-hour active budget. Routing stays dense for 28.75\% of that budget and reaches hard Top-2 at 81.25\%. At that point, each policy step evaluates two of the group's 32 specialists.

\noindent\begin{minipage}{\linewidth}
\centering
\captionof{table}{Stage~2 catalog and rollout coverage at the final checkpoints. Hours per expert reports catalog hours divided by the specialist count.}
\label{tab:corpus}
\begingroup
\small
\setlength{\tabcolsep}{3.5pt}
\begin{tabularx}{\linewidth}{Lrrrrrr}
\toprule
Motion source & Groups & Experts & Catalog (h) & h / Expert & Visits (M) & Distinct frames \\
\midrule
LAFAN1, Teleoperation, AMASS & 1 & 32 & 68.1 & 2.13 & 72.2 & 1.55M (12.7\%) \\
BONES-SEED & 2 & 64 & 263.2 & 4.11 & 139.2 & 5.58M (11.8\%) \\
ChingMu & 5 & 160 & 1,113.3 & 6.96 & 315.6 & 13.10M (6.5\%) \\
MotionMillion & 8 & 256 & 1,831.9 & 7.16 & 544.9 & 19.68M (6.0\%) \\
OmniHuMo & 16 & 512 & 6,826.9 & 13.33 & 1,084.2 & 40.69M (3.3\%) \\
\midrule
Total & 32 & 1,024 & 10,103.5 & 9.87 & 2,156.0 & 80.61M (4.4\%) \\
\bottomrule
\end{tabularx}
\endgroup

\smallskip
{\footnotesize\raggedright Across groups, 183.68M distinct frames were read on the CPU, 98.32M reached GPU pages (including lookahead), and 80.61M supplied rollouts (448~hours at 50~Hz). Each group has 262,144 GPU slots. Visits count replay-admitted transitions with repeats. Coverage divides distinct rollout frames by frames eligible for sampling (1.817B in clips of at least 44 frames).\par}
\end{minipage}

\noindent\begin{minipage}{\linewidth}
\centering
\begingroup
\definecolor{routingink}{HTML}{173047}
\definecolor{routingmuted}{HTML}{536879}
\definecolor{routinggrid}{HTML}{DCE5EE}
\definecolor{routingselected}{HTML}{174C91}
\definecolor{routingzero}{HTML}{F0F4F7}

\newcommand{\routingshade}[1]{%
  \pgfmathsetmacro{\shade}{min(100,100*sqrt(#1/0.8))}%
  \colorlet{routingheat}{routingselected!\shade!routingzero}%
}

\newcommand{\routingpanel}[5]{%
  \begin{scope}[shift={(#1,0)}]
    \draw[frame] (0,-0.35) rectangle (5.20,-4.12);
    \node[panel title] at (2.60,-0.68) {#2};
    \node[note] at (2.60,-1.00) {#3};
    \foreach \level in {0,0.2,0.4,0.6,0.8}{
      \draw[routinggrid,line width=0.45pt] (0.70,{-3.22+2.43*\level}) -- (5.01,{-3.22+2.43*\level});
      \node[tick,anchor=east] at (0.57,{-3.22+2.43*\level})
        {\pgfmathprintnumber[fixed,precision=1]{\level}};
    }
    \foreach \weight [count=\expert] in {#4}{
      \pgfmathsetmacro{\barcenter}{0.785+(\expert-1)*0.132}
      \routingshade{\weight}
      \fill[routingheat] ({\barcenter-0.053},-3.22) rectangle ({\barcenter+0.053},{-3.22+2.43*\weight});
    }
    \foreach \expert/\tone in {1/routingmuted,7/routingselected,16/routingmuted,25/routingselected,32/routingmuted}{
      \node[tick,text=\tone] at ({0.785+(\expert-1)*0.132},-3.43) {\expert};
    }
    \node[note] at (2.60,-3.80) {#5};
  \end{scope}%
}
\newcommand{\routingheatcol}[2]{%
  \pgfmathsetmacro{\colleft}{2.426+11.14*#1/32}
  \foreach \weight [count=\expert] in {#2}{
    \pgfmathsetmacro{\rowtop}{-5.00-(\expert-1)*0.0675}
    \routingshade{\weight}
    \fill[routingheat]
      (\colleft,\rowtop) rectangle ({\colleft+0.3482},{\rowtop-0.059});
  }%
}
\begin{tikzpicture}[
  x=1cm,y=1cm,
  every node/.style={font=\reportfont\fontsize{8.2}{10}\selectfont,text=routingink,inner sep=0pt},
  panel title/.style={font=\reportfont\bfseries\fontsize{9.4}{11}\selectfont},
  note/.style={text=routingmuted},
  tick/.style={font=\reportfont\fontsize{7.2}{8.5}\selectfont,text=routingmuted},
  frame/.style={rounded corners=3pt,draw=routinggrid,line width=0.55pt},
  progress arrow/.style={draw=routingmuted,line width=0.65pt,-{Latex[length=2.65pt,width=2.5pt]}}
]
  \path[use as bounding box] (0,0.22) rectangle (16.34,-7.77);
  \node[note,anchor=east] at (16.34,0) {Fixed reference logits};
  \begin{scope}[yshift=5.05cm,yscale=1.13]
  \node[panel title,anchor=west] at (0,-4.47) {Routing weights during annealing};
  \draw[frame] (0,-4.72) rectangle (16.34,-7.76);
  \routingheatcol{0}{%
  0.00671,0.0170066,0.0410013,0.004776,0.0136481,0.0570314,0.250536,0.0097143,%
  0.0031381,0.0236557,0.0842343,0.0060111,0.0374723,0.0038713,0.0149334,0.028321,%
  0.0717798,0.0087024,0.0025183,0.0184231,0.0118651,0.0500791,0.0055489,0.0216197,%
  0.0921671,0.0105234,0.0345913,0.0043649,0.0074157,0.0251185,0.0022336,0.0309881}
  \routingheatcol{1}{%
  0.0049263,0.0144847,0.0389077,0.0032905,0.0112597,0.0559246,0.2939193,0.0075991,%
  0.0019846,0.0210566,0.0854168,0.0043258,0.0352169,0.002558,0.0124847,0.0257762,%
  0.0718414,0.0066845,0.0015179,0.0158662,0.0095822,0.0485003,0.0039342,0.0190224,%
  0.102517,0.0083394,0.0322228,0.0029546,0.0055418,0.0225282,0.0013102,0.0285054}
  \routingheatcol{2}{%
  0.0033616,0.0119757,0.0365516,0.0020576,0.008947,0.0544831,0.3371496,0.0056407,%
  0.0010981,0.0183658,0.0862405,0.0028727,0.0327241,0.0015078,0.0100869,0.0230809,%
  0.0715446,0.0048446,0.0007824,0.0132978,0.0074111,0.0466107,0.0025599,0.0163626,%
  0.1125142,0.006295,0.0296405,0.0018023,0.0038727,0.0198266,0.0006483,0.0258433}
  \routingheatcol{3}{%
  0.002063,0.0095282,0.0339312,0.0011107,0.006764,0.0526759,0.379654,0.003894,%
  0.0004902,0.015618,0.0866257,0.0016951,0.0299991,0.0007437,0.0077922,0.0202595,%
  0.0708325,0.0032365,0.000311,0.0107637,0.005407,0.0443917,0.0014661,0.0136796,%
  0.1220638,0.0044458,0.0268557,0.000937,0.0024587,0.0170452,0.0002406,0.0230204}
  \routingheatcol{4}{%
  0.0010726,0.0072015,0.0310541,0.0004689,0.0047733,0.0504799,0.4208495,0.0024174,%
  0.0001475,0.0128591,0.0864983,0.0008283,0.0270569,0.0002686,0.0056623,0.0173472,%
  0.0696546,0.0019148,0.0000748,0.0083206,0.0036323,0.0418331,0.0006827,0.0110238,%
  0.1310756,0.0028521,0.0238895,0.000371,0.0013471,0.0142263,0.0000503,0.0200659}
  \routingheatcol{5}{%
  0.0004167,0.0050661,0.0279419,0.0001221,0.0030453,0.0478864,0.4602065,0.0012682,%
  0.0000176,0.0101487,0.0858007,0.0002872,0.0239263,0.0000498,0.0037687,0.0143931,%
  0.0679763,0.000929,0.0000047,0.0060379,0.0021539,0.0389389,0.000216,0.0084589,%
  0.1394766,0.0015758,0.0207769,0.0000844,0.000574,0.0114266,0.000002,0.0170228}
  \routingheatcol{6}{%
  0.0000862,0.0032043,0.0246351,0.0000083,0.0016555,0.0449104,0.4973267,0.0004912,%
  0,0.0075626,0.0845091,0.000045,0.0206547,0.0000009,0.0021907,0.0114637,%
  0.0657918,0.0003099,0,0.0039977,0.0010375,0.0357344,0.0000266,0.0060641,%
  0.1472306,0.000671,0.0175711,0.0000036,0.0001463,0.008719,0,0.0139518}
  \routingheatcol{7}{%
  0.0000018,0.0017054,0.0211937,0,0.0006737,0.0415917,0.5319753,0.0000953,%
  0,0.0051922,0.082638,0.0000002,0.0173079,0,0.0010081,0.0086424,%
  0.0631272,0.00004,0,0.0022925,0.0003321,0.032267,0,0.003933,%
  0.1543443,0.0001637,0.0143431,0,0.0000076,0.0061928,0,0.0109311}
  \routingheatcol{8}{%
  0,0.0006547,0.0176888,0,0.00014,0.0379828,0.5640276,0.0000008,%
  0,0.0031407,0.0802228,0,0.0139631,0,0.0002831,0.0060254,%
  0.0600253,0,0,0.0010168,0.0000337,0.0285957,0,0.0021685,%
  0.160848,0.0000054,0.0111759,0,0,0.0039497,0,0.0080513}
  \routingheatcol{9}{%
  0,0.0001081,0.0141966,0,0.000001,0.0341371,0.5934041,0,%
  0,0.0015209,0.0773,0,0.0107048,0,0.0000142,0.0037211,%
  0.0565293,0,0,0.0002512,0,0.0247813,0,0.0008784,%
  0.1667733,0,0.0081615,0,0,0.0021027,0,0.0054143}
  \routingheatcol{10}{%
  0,0,0.0108009,0,0,0.0301076,0.6200553,0,%
  0,0.0004509,0.0739032,0,0.0076281,0,0,0.0018547,%
  0.0526802,0,0,0.000004,0,0.0208867,0,0.0001591,%
  0.1721491,0,0.0054053,0,0,0.0007767,0,0.0031381}
  \routingheatcol{11}{%
  0,0,0.007599,0,0,0.0259483,0.6439601,0,%
  0,0.0000178,0.0700638,0,0.0048464,0,0,0.0005723,%
  0.0485172,0,0,0,0,0.0169792,0,0,%
  0.1770023,0,0.0030349,0,0,0.000093,0,0.0013657}
  \routingheatcol{12}{%
  0,0,0.0047123,0,0,0.0217206,0.665146,0,%
  0,0,0.0658188,0,0.0025048,0,0,0.0000215,%
  0.0440846,0,0,0,0,0.0131378,0,0,%
  0.181366,0,0.0012146,0,0,0,0,0.0002729}
  \routingheatcol{13}{%
  0,0,0.0023063,0,0,0.0175069,0.6837579,0,%
  0,0,0.0612363,0,0.0008028,0,0,0,%
  0.0394512,0,0,0,0,0.0094662,0,0,%
  0.1853087,0,0.0001637,0,0,0,0,0}
  \routingheatcol{14}{%
  0,0,0.0006123,0,0,0.0133938,0.6999354,0,%
  0,0,0.0563784,0,0.0000194,0,0,0,%
  0.0346815,0,0,0,0,0.0060871,0,0,%
  0.1888922,0,0,0,0,0,0,0}
  \routingheatcol{15}{%
  0,0,0,0,0,0.0094946,0.713917,0,%
  0,0,0.0513385,0,0,0,0,0,%
  0.0298646,0,0,0,0,0.0031708,0,0,%
  0.1922144,0,0,0,0,0,0,0}
  \routingheatcol{16}{%
  0,0,0,0,0,0.0059692,0.7261532,0,%
  0,0,0.046288,0,0,0,0,0,%
  0.025147,0,0,0,0,0.0009796,0,0,%
  0.1954631,0,0,0,0,0,0,0}
  \routingheatcol{17}{%
  0,0,0,0,0,0.0029503,0.7366208,0,%
  0,0,0.0412357,0,0,0,0,0,%
  0.020553,0,0,0,0,0,0,0,%
  0.1986402,0,0,0,0,0,0,0}
  \routingheatcol{18}{%
  0,0,0,0,0,0.0006983,0.7452805,0,%
  0,0,0.0361804,0,0,0,0,0,%
  0.0161054,0,0,0,0,0,0,0,%
  0.2017354,0,0,0,0,0,0,0}
  \routingheatcol{19}{%
  0,0,0,0,0,0,0.7522005,0,%
  0,0,0.031158,0,0,0,0,0,%
  0.0118618,0,0,0,0,0,0,0,%
  0.2047797,0,0,0,0,0,0,0}
  \routingheatcol{20}{%
  0,0,0,0,0,0,0.7577542,0,%
  0,0,0.026327,0,0,0,0,0,%
  0.0079724,0,0,0,0,0,0,0,%
  0.2079463,0,0,0,0,0,0,0}
  \routingheatcol{21}{%
  0,0,0,0,0,0,0.7622171,0,%
  0,0,0.0218222,0,0,0,0,0,%
  0.0045783,0,0,0,0,0,0,0,%
  0.2113824,0,0,0,0,0,0,0}
  \routingheatcol{22}{%
  0,0,0,0,0,0,0.765508,0,%
  0,0,0.017641,0,0,0,0,0,%
  0.0017768,0,0,0,0,0,0,0,%
  0.2150743,0,0,0,0,0,0,0}
  \routingheatcol{23}{%
  0,0,0,0,0,0,0.7673808,0,%
  0,0,0.0137044,0,0,0,0,0,%
  0,0,0,0,0,0,0,0,%
  0.2189148,0,0,0,0,0,0,0}
  \routingheatcol{24}{%
  0,0,0,0,0,0,0.767451,0,%
  0,0,0.0098546,0,0,0,0,0,%
  0,0,0,0,0,0,0,0,%
  0.2226944,0,0,0,0,0,0,0}
  \routingheatcol{25}{%
  0,0,0,0,0,0,0.7665935,0,%
  0,0,0.0065133,0,0,0,0,0,%
  0,0,0,0,0,0,0,0,%
  0.2268932,0,0,0,0,0,0,0}
  \routingheatcol{26}{%
  0,0,0,0,0,0,0.7647654,0,%
  0,0,0.003711,0,0,0,0,0,%
  0,0,0,0,0,0,0,0,%
  0.2315236,0,0,0,0,0,0,0}
  \routingheatcol{27}{%
  0,0,0,0,0,0,0.761919,0,%
  0,0,0.0014869,0,0,0,0,0,%
  0,0,0,0,0,0,0,0,%
  0.2365941,0,0,0,0,0,0,0}
  \routingheatcol{28}{%
  0,0,0,0,0,0,0.7579373,0,%
  0,0,0,0,0,0,0,0,%
  0,0,0,0,0,0,0,0,%
  0.2420627,0,0,0,0,0,0,0}
  \routingheatcol{29}{%
  0,0,0,0,0,0,0.752375,0,%
  0,0,0,0,0,0,0,0,%
  0,0,0,0,0,0,0,0,%
  0.247625,0,0,0,0,0,0,0}
  \routingheatcol{30}{%
  0,0,0,0,0,0,0.7460814,0,%
  0,0,0,0,0,0,0,0,%
  0,0,0,0,0,0,0,0,%
  0.2539186,0,0,0,0,0,0,0}
  \routingheatcol{31}{%
  0,0,0,0,0,0,0.7390001,0,%
  0,0,0,0,0,0,0,0,%
  0,0,0,0,0,0,0,0,%
  0.2609999,0,0,0,0,0,0,0}
  \routingheatcol{32}{%
  0,0,0,0,0,0,0.7310586,0,%
  0,0,0,0,0,0,0,0,%
  0,0,0,0,0,0,0,0,%
  0.2689414,0,0,0,0,0,0,0}
  \node[note,rotate=90] at (1.34,-6.08) {Expert ID};
  \foreach \expert/\tone in {1/routingmuted,7/routingselected,16/routingmuted,25/routingselected,32/routingmuted}{
    \pgfmathsetmacro{\rowcenter}{-5.03-(\expert-1)*0.0675}
    \node[tick,anchor=east,text=\tone] at (2.26,\rowcenter) {\expert};
  }
  \foreach \progress in {0,0.25,0.5,0.75,1}{
    \pgfmathsetmacro{\tickx}{2.60+11.14*\progress}
    \draw[routingmuted,line width=0.35pt] (\tickx,-7.18) -- (\tickx,-7.25);
    \node[tick] at (\tickx,-7.37) {\progress};
  }
  \node[note] at (8.17,-7.62) {Routing progress \(s\)};
  \node[note,anchor=west] at (14.59,-5.07) {Weight};
  \foreach \index in {0,...,63}{
    \pgfmathsetmacro{\weight}{0.8*\index/63}
    \routingshade{\weight}
    \fill[routingheat] (14.61,{-7.12+1.82*\index/64})
      rectangle (15.03,{-7.12+1.82*(\index+1)/64});
  }
  \foreach \weight in {0,0.2,0.4,0.6,0.8}{
    \pgfmathsetmacro{\legendy}{-7.12+1.82*\weight/0.8}
    \draw[routingmuted,line width=0.35pt] (15.03,\legendy) -- (15.10,\legendy);
    \node[tick,anchor=west] at (15.19,\legendy) {\weight};
  }
  \foreach \sample in {0,0.5,1}{
    \pgfmathsetmacro{\slicex}{2.60+11.14*\sample}
    \draw[routingselected,line width=0.55pt]
      ({\slicex-0.174},-4.98) rectangle ({\slicex+0.174},-7.17);
  }
  \end{scope}
  \foreach \sample in {0,0.5,1}{
    \pgfmathsetmacro{\slicex}{2.60+11.14*\sample}
    \draw[progress arrow,routingselected,line width=0.55pt]
      (\slicex,-3.77) -- (\slicex,-4.25);
  }
  \begin{scope}[yshift=-3.98cm,yscale=0.91]
  \routingpanel{0}{Dense softmax}{\(s=0,\ \alpha=1\)}
      {%
  0.00671,0.0170066,0.0410013,0.004776,0.0136481,0.0570314,0.250536,0.0097143,%
  0.0031381,0.0236557,0.0842343,0.0060111,0.0374723,0.0038713,0.0149334,0.028321,%
  0.0717798,0.0087024,0.0025183,0.0184231,0.0118651,0.0500791,0.0055489,0.0216197,%
  0.0921671,0.0105234,0.0345913,0.0043649,0.0074157,0.0251185,0.0022336,0.0309881}
      {32 nonzero weights}
  \routingpanel{5.57}{Entmax blend}{\(s=0.5,\ \alpha=1.5\)}
      {%
  0,0,0,0,0,0.0059692,0.7261532,0,%
  0,0,0.046288,0,0,0,0,0,%
  0.025147,0,0,0,0,0.0009796,0,0,%
  0.1954631,0,0,0,0,0,0,0}
      {6 nonzero weights}
  \routingpanel{11.14}{Hard Top-2}{\(s=1,\ \alpha=2\)}
      {%
  0,0,0,0,0,0,0.7310586,0,%
  0,0,0,0,0,0,0,0,%
  0,0,0,0,0,0,0,0,%
  0.2689414,0,0,0,0,0,0,0}
      {2 nonzero weights (7 and 25)}
  \end{scope}
\end{tikzpicture}
\endgroup


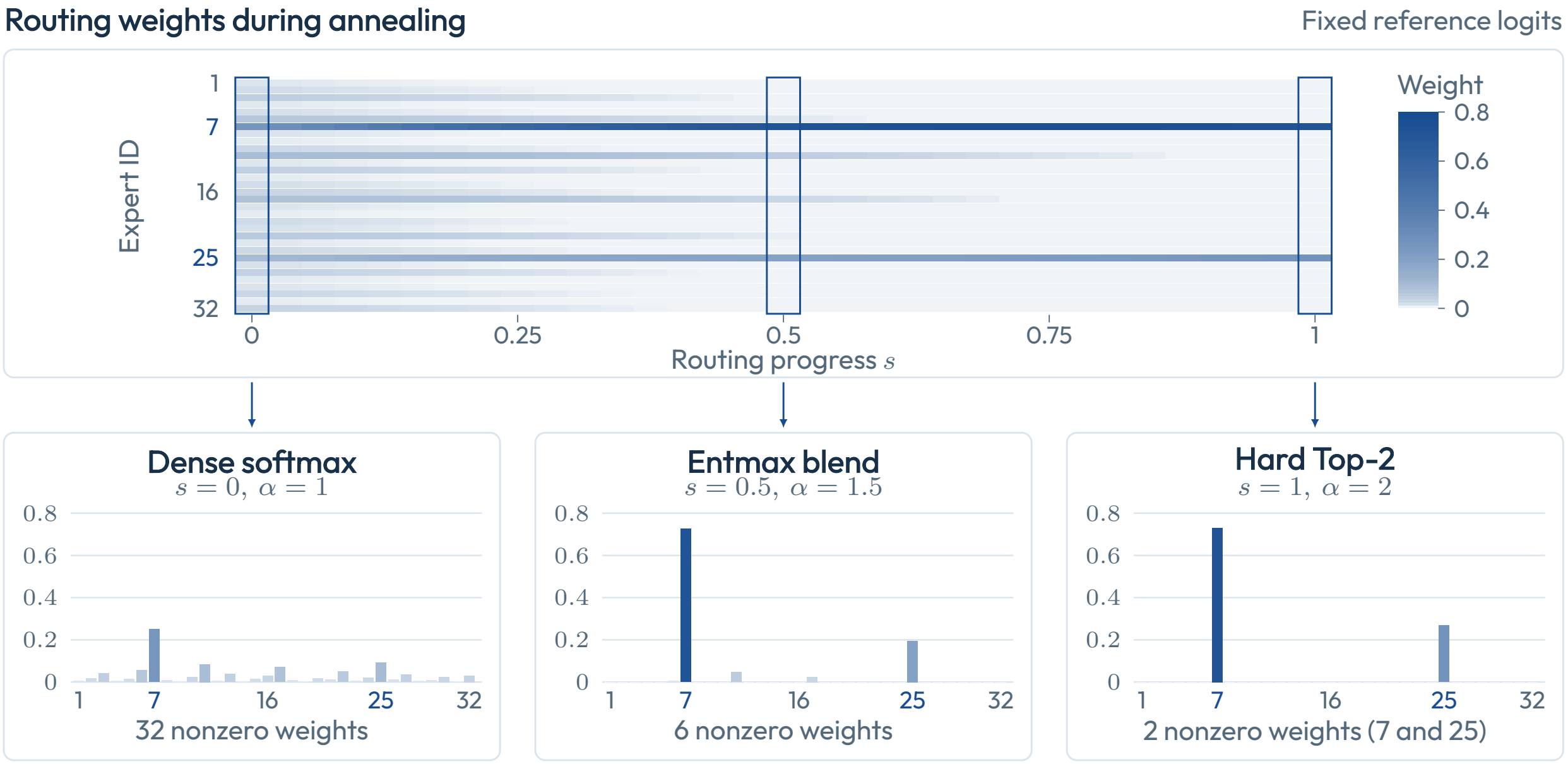
\captionof{figure}{Annealed Top-\(K\) Routing with 32 experts and \(K=2\). The heatmap shows the full transition at fixed logits and shares its weight scale with the snapshots. The marked columns at \(s=0,0.5,1\) are expanded below. The blend at \(s=0.5\) retains six experts. At \(s=1\), the two selected experts receive weights of 0.73 and 0.27. During training, the logits and selected experts can change.}
\label{fig:annealed-routing}
\end{minipage}

\subsection{Global routing and residual gating}
Global routing selects corrections across groups, and the gate controls their strength. We stack the local \(E\times144\) router matrices row-wise and assemble the specialists in the same group order. The reference feature layout and normalization statistics are shared. We verify each group's original logits, then mask the other groups and restore its gain. The resulting 29-dimensional actions match within \(2.4\times10^{-6}\) on CPU. Stage~1 supplies the critic and dynamics.

Distillation updates only the global router. Group labels select local teachers \(q_K^{(j)}(r)\) whose Top-\(K\) indices map into the global bank. The reference-only student minimizes \(-\sum_i q^{(j)}_{K,i}(r)\log\operatorname{softmax}(z_{\mathrm{global}}(r))_i\) over all \(N\cdot E\) experts. We take four vectors per clip from 12 clips per group. Each of 300 Adam updates at \(10^{-3}\) draws 16 vectors per group. On these 1,536 calibration inputs, the global Top-2 recovers 171 of 3,072 teacher selections before fitting and 2,068 after (5.6\% to 67.3\%).

We train the global router for one active hour with equal reset probabilities across the seven libraries. The residual scale rises from zero to one. The router stays dense for the first 7.5\% and runs at hard Top-2 for the final 17.5\% of the active budget. We then freeze the router and fit the gate for two active hours on six libraries. LAFAN1 receives reset probability \(1/3\) and the other five receive \(2/15\) each. OmniHuMo experts remain in the bank. The generalist and all specialists stay frozen throughout. Rollouts still evaluate the selected networks, and the critic and dynamics continue learning.

\begin{samepage}
\paragraph{Learned residual strength.}
We let the reference set the gain so that the generalist can act on its own when a correction is unhelpful. The residual block takes the actor's hidden state \(x\) and reference features \(r\). The gate reads \(r\) alone.
\begin{equation}
h(x,r)=f_{\mathrm{shared}}(x)+g(r)\sum_{i\in T_K(r)}w_i(r)f_i(x),
\qquad
g(r)=\operatorname{clip}_{[0,1]}\!\left(\frac{a^\top r+b+3}{6}\right).
\label{eq:gate}
\end{equation}
\end{samepage}
Here \(T_K(r)\) holds the selected experts and \(w_i\) are their normalized weights. Gate fitting updates \(a,b\) and preserves all inherited actor tensors bitwise. At fixed references, changing robot state and history leaves router and gate logits bitwise identical. When \(g(r)=0\), the block returns the shared output.

\newgeometry{margin=0.62in}
\begin{landscape}
\thispagestyle{empty}
\noindent\begin{minipage}{\linewidth}
\noindent{\reportfont\bfseries\Large Training reference motions}\hfill
{\reportfont\small First frame\enspace
\tikz[baseline=-0.52ex]{
  \definecolor{poseone}{HTML}{BFC6EA}
  \definecolor{posetwo}{HTML}{BDD3EF}
  \definecolor{posethree}{HTML}{A9D7E2}
  \definecolor{posefour}{HTML}{9BCDD0}
  \definecolor{posefive}{HTML}{34414D}
  \foreach \tone [count=\i from 0] in {poseone,posetwo,posethree,posefour,posefive}
    \fill[\tone] (\i*0.20,0) circle[radius=0.072cm];
  \draw[black!50,-{Latex[length=2.7pt]},line width=0.4pt] (0.94,0) -- (1.23,0);
}\enspace Final frame}\hfill
{\reportfont\small 32 Stage 2 groups \enspace\textbar\enspace 128 reference clips}\par
\vspace{0.26cm}
\centering
\includegraphics[width=\linewidth]{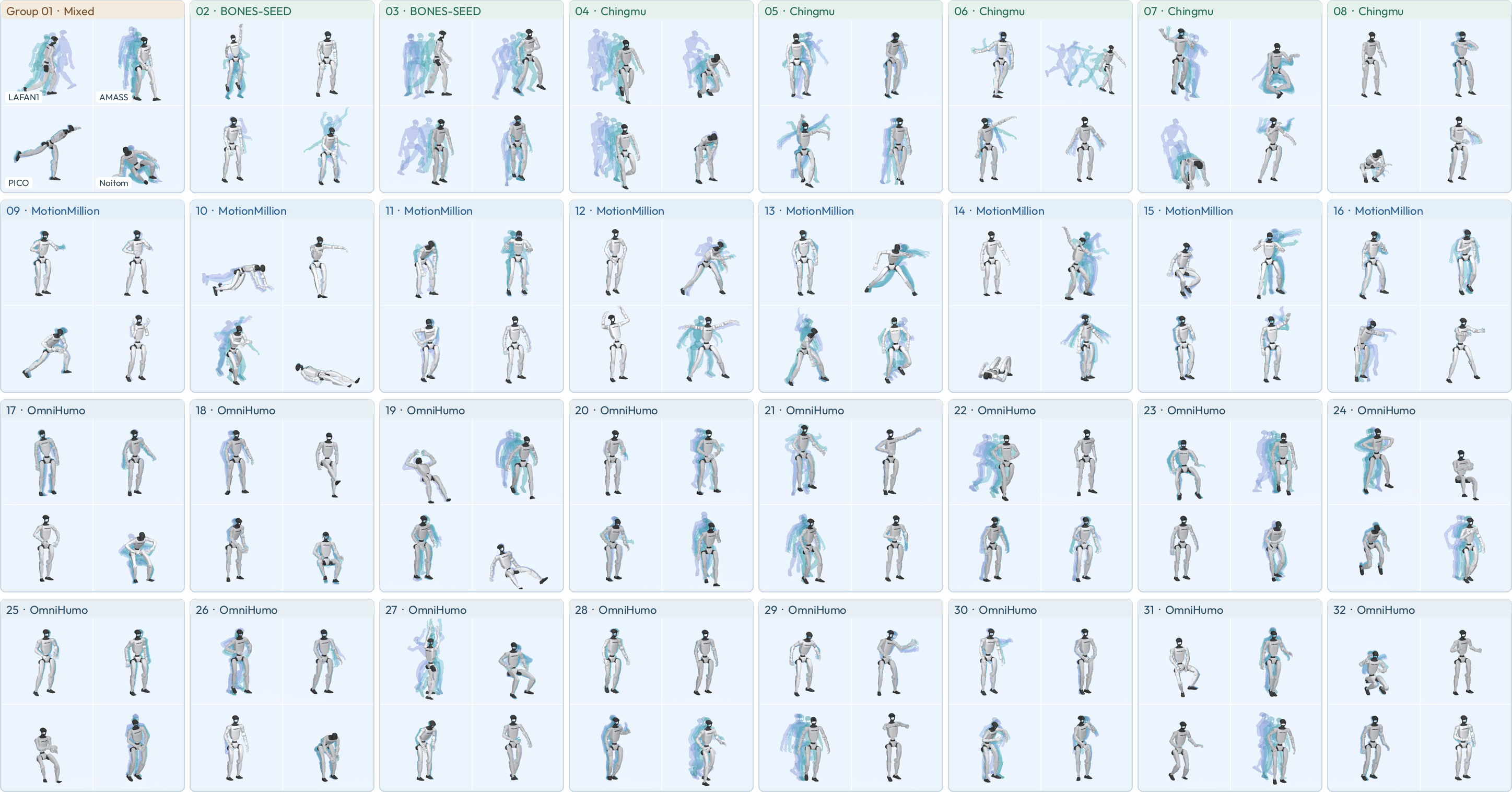}
\captionof{figure}{Reference motions for generalist and specialist learning. Groups 01--16 illustrate the six sources used for the Stage~1 generalist. Stage~2 adds OmniHuMo (groups 17--32) and applies the group partition shown here. Each tile shows four Stage~2 clips rendered on the G1. Five poses span up to 0.96 seconds, with a nominal 0.24-second interval shortened near clip boundaries. Four translucent poses lead to a solid final pose. Group 01 shows LAFAN1, AMASS, and in-house PICO and Noitom recordings. Groups 02--03 show BONES-SEED, 04--08 ChingMu, and 09--16 MotionMillion.}
\label{fig:stage2-motion-gallery}
\end{minipage}\par
\vfill\noindent\makebox[\linewidth]{\small\thepage}\par
\end{landscape}
\restoregeometry

\newgeometry{hmargin=1in,top=0.86in,bottom=0.74in}

\subsection{Implementation details}
\label{sec:implementation}

We train one real-time actor across both benchmark G1 setups, with 2,048 parallel environments per setup. Each 50~Hz control step uses four 5~ms explicit-PD substeps.

\paragraph{Actor and routing.}
The actor maps its eight most recent observation frames to 29 joint targets with a 512-wide causal block, eight query heads, and four key/value heads. The 406.1M parameters occupy 1.62~GB in FP32. Each control step uses 4.23M parameters: 3.30M for the shared path, 147,456 for the router scoring all 1,024 experts, 145 for the gate, and 393,216 for each of the two selected width-256 expert MLPs. Evaluation keeps the full bank resident. Before attention, the router and gate read 36 values per reference pose at four future offsets (20, 120, 220, and 320~ms): root-frame displacement, joints, height, and projected gravity.

\section{Scalable Motion Data and Training Infrastructure}

\begin{figure}[!ht]
\centering
\includegraphics[width=\linewidth]{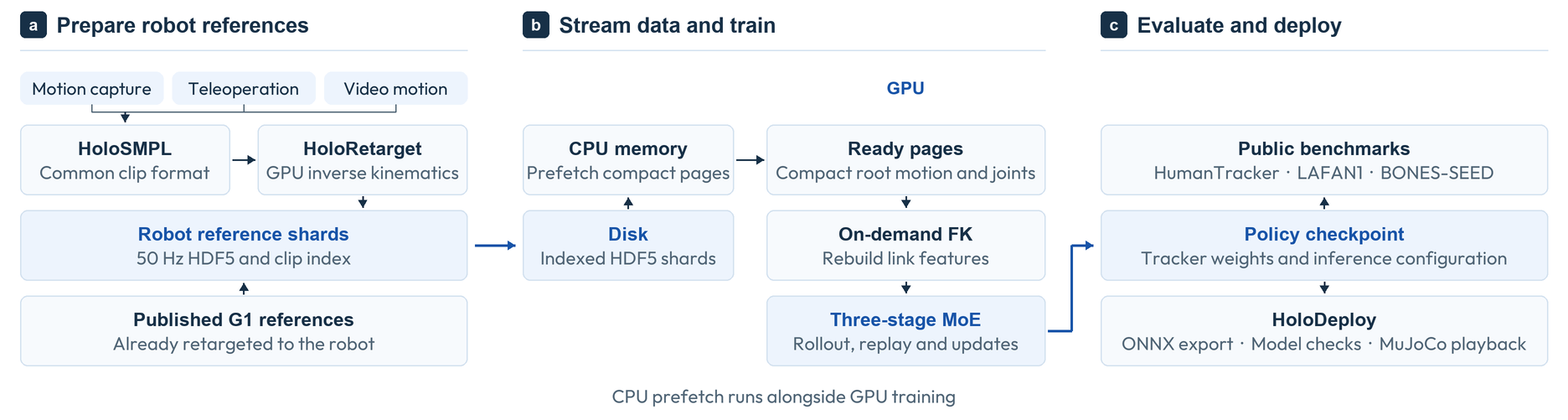}
\caption{HoloMotion data and training infrastructure. (a) HoloSMPL and HoloRetarget prepare robot references from heterogeneous sources. Published G1 trajectories enter directly. (b) CPU workers prepare compact pages while the GPU reconstructs link states as needed and runs training updates. (c) The saved policy feeds benchmark evaluation and HoloDeploy separately.}
\label{fig:workflow}
\end{figure}

Figure~\ref{fig:workflow} shows the workflow from motion preparation to deployment.

\subsection{Motion catalog and retargeting}
\label{sec:retargeting}

We turn motions from different sources into references for the same tracking controller. HoloSMPL standardizes the human clips, and HoloRetarget maps them to the target robot.

\begin{figure}[!t]
\centering
\input{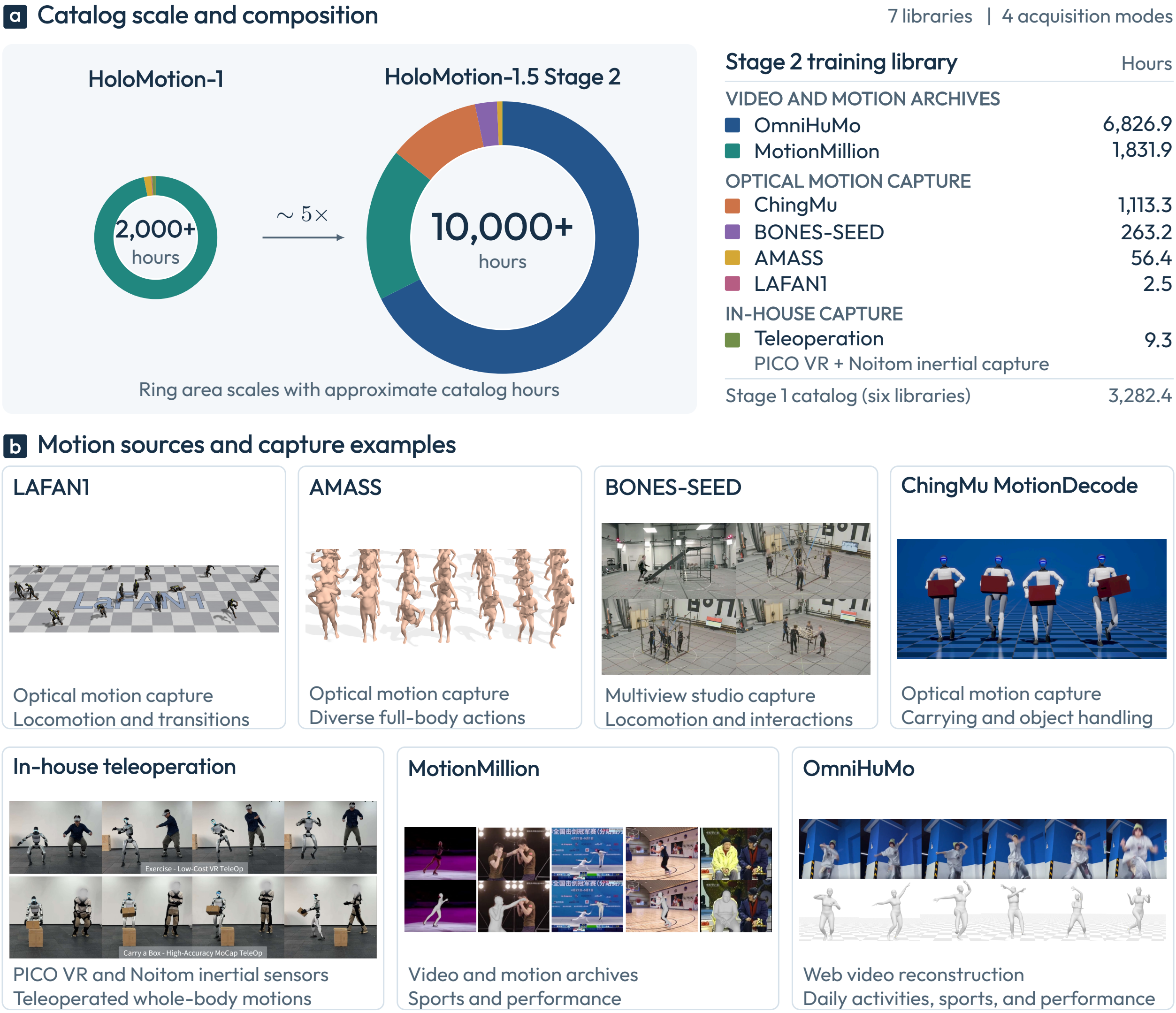}
\caption{Motion catalogs and sources. (a) Catalog hours from HoloMotion-1 Table~4~\citep{holomotion1} and our prepared 50~Hz references. (b) Images from the LAFAN1, BONES-SEED, and MotionDecode releases~\citep{lafan1,bonesseed,chingmudata}, AMASS Fig.~1~\citep{amass}, HoloMotion-1 Fig.~2~\citep{holomotion1}, MotionMillion Fig.~3b~\citep{motionmillion}, and AnyMo Fig.~S3~\citep{omnihumo}.}
\label{fig:training-motions}

\end{figure}

\paragraph{HoloSMPL.}
HoloSMPL gives diverse motion sources one format for validation and retargeting. Our inputs span optical motion capture (LAFAN1, AMASS, BONES-SEED, and ChingMu's MotionDecode)~\citep{lafan1,amass,bonesseed,chingmudata}, in-house teleoperation with PICO 4 Ultra and Noitom PN Link~\citep{holomotion1}, and video-derived motions from MotionMillion and OmniHuMo~\citep{motionmillion,omnihumo}. MotionMillion also includes archived captures. Source adapters map the inputs to SMPL body parameters with a common joint layout, Z-up coordinates in meters, and 50~Hz timing~\citep{smpl,holosmpl}. Indexed HDF5 shards preserve the root motion, joint rotations, one body-shape vector per clip, and source metadata. The six-library generalist catalog contains 3,282.4 hours. Stage~2 adds 6,826.9 hours from 4.18 million prepared OmniHuMo clips and takes the published LAFAN1 and BONES-SEED G1 trajectories directly as robot references. The resulting catalog contains 10,103.5 hours (Figure~\ref{fig:training-motions}).

\paragraph{HoloRetarget.}
HoloRetarget maps 24 body poses to the G1 root and 29 joints for both offline data and live PICO teleoperation~\citep{holoretarget}. It retains GMR's scales and offsets~\citep{gmr}, retunes task weights, and adds hand targets. It seeds the base from the pelvis and reuses the preceding joint solution. Newton and Warp replay a damped IK step with analytic Jacobians and joint limits. SOMA Retargeter uses the same GPU libraries for a different skeleton~\citep{somaretgt}. On one cluster RTX 5090, the warmed-up offline path converts 2,115 SMPL frames/s, or 2,104 with local HDF5 writes. Both are medians of three trials on one AMASS walking-and-turning clip.

\textbf{HoloDeploy.} The Rust control loop runs in MuJoCo and on hardware, with a separate safety process on the robot. The ONNX policy and its configuration support model checks and replay. On Jetson Orin NX, Rust/CPU cut retargeting IK p99 from 3.91 to 0.038~ms relative to Python/GPU in a 540-frame offline replay. In 500-step checks of the HoloMotion v1.4.1 policy, median policy-step latency was 1.25~ms, and p99 fell from 2.88--3.19~ms in the earlier image to 1.32~ms. These gains reduce the time spent on retargeting and policy execution within the control period.

\subsection{Compact references and asynchronous streaming}
\label{sec:infrastructure}

Resets, simulation, and replay repeatedly query the same motion. To avoid host transfers on each read, the simulator and learner gather from GPU pages. Stored frames contain the root pose and joint positions. Training pages add velocities and omit derived link states to fit more clips in GPU memory. We reconstruct link states for the reward with batched GPU forward kinematics, reused scratch buffers, tensor-core rotation products, and CUDA Graph replay. The actor reads root and joint targets directly, and we cache frequently queried link positions for the critic~\citep{mjlabloader}.

\paragraph{Asynchronous loading.}
Preloading the catalog delays the first rollout, and overwriting old pages can invalidate references that active episodes or replay will read again. We start training on the first ready pages while a bounded CPU queue prepares new pages for transfer into alternating GPU regions. New episodes sample only fully published clips. When a region retires, it stops admitting resets but remains readable until active rollouts and replay release their references. By separating reset eligibility from the lifetime of stored frames, we can rotate through a larger catalog while training continues on ready clips.

\paragraph{Reference-streaming measurements.}
A stored pose has 36 FP32 values. Training pages use 71 values (284 bytes per frame), compared with 1,336 bytes for the previous representation. The 317-clip teleoperation bank shrank from 2.26 to 0.86~GB, including auxiliary tensors. Compact and native storage matched on 240 fixed-LAFAN1 trials across three seeds. In a same-seed test, time to first rollout fell from 34.6 to 7.5 minutes; storage cache state was uncontrolled. A separate fixed-step test measured 2.91M versus 31.76M distinct rollout frames, with 4.9\% more active training time for the rotating bank.

\subsection{Learner and simulator execution}
\label{sec:execution}
\enlargethispage{12pt}
Repeated history encoding dominated early actor updates. We compute history keys and values once, reuse them across imagined steps with gradients intact, and apply single-query attention to new tokens. With compiled, full-batch rollouts, early profiling measured 11.9 to 0.44 seconds per update (\(27\times\)).

We also batch dynamics gradients across time and use activation checkpointing to reduce rollout memory~\citep{xmimic}. CUDA Graph replay cuts launch overhead in the simulator and learner~\citep{instanttrack}. 

\section{Evaluation}

We test one checkpoint for held-out transfer on HumanTracker and tracking fidelity on MotionChallenge, extending HoloMotion-1's in-house evaluation~\citep{holomotion1,humantracker,motionchallenge,motionchallengecore}. Tables~\ref{tab:full-humantracker}--\ref{tab:full-seed} compare results published through 27 November 2026.

\paragraph{HumanTracker.}
HumanTracker tests transfer across motions with different demands on balance and contact. The source corpus contains about 153 hours of optical recordings from 24 performers. Its test split contains 974 daily motions, 268 highly dynamic motions, 1,094 interactions, and 164 ground motions. These families probe steady locomotion, rapid changes in support, coordinated gestures, and transitions involving low postures or multiple contacts. We reserve all 2,500 test clips and use the benchmark's GMR references, G1 scene in MuJoCo 3.10.0, actuator limits, and whole-body termination~\citep{humantracker, humantrackerdata, humantrackerreleased}. We rank by the fraction of clips completed without failure, giving each clip equal weight. Completion measures sustained tracking, joint-angle MPJPE measures pose accuracy, and HumanScore estimates perceived motion quality from human preferences. The evaluator weights MPJPE by reference frames and HumanScore by scored frames (Table~\ref{tab:full-humantracker}).

\paragraph{MotionChallenge.}
MotionChallenge measures tracking fidelity on shared robot references and publishes the rankings online. Its LAFAN1 track uses 40 animation clips, while BONES-SEED tests tracking across 129,561 references. Both tracks use Core's G1 model in MJWarp 3.11.0 at 50~Hz. The actor follows the track's observation rules, which exclude global position and actual or estimated linear velocity~\citep{motionchallengecore,motionchallengeprotocol}. Both sets are in the training catalog~\citep{lafanretarget,bonesseed}. The 44-frame training minimum excludes 26 BONES-SEED clips, which remain in evaluation.

Ranking uses \emph{No termination / XY only} mean per-keypoint position error (MPKPE). XY alignment removes the horizontal pelvis offset at each frame, retaining height and heading errors. The score averages 3D distances across 14 G1 body origins and all post-reset frames, including after failure, then weights motions equally. The prefix includes the first sample with pelvis-height error over 0.25~m or pelvis-frame gravity-Z error over 0.8. Success requires reaching the endpoint without failure~\citep{motionchallengeprotocol}.

\newgeometry{hmargin=0.73in,vmargin=1in}
\begin{landscape}
\definecolor{benchmarklead}{HTML}{E7F0FD}
\thispagestyle{empty}
\subsection{HumanTracker results}

The held-out split tests how well the final controller transfers to unfamiliar motions. HoloMotion-1.5 completes \textbf{2,259 of 2,500} held-out motions, 203 more than HoloMotion v1.4.0 and 20 more than the count reconstructed for SONIC~1.1. It has the lowest directly measured overall joint-angle MPJPE among ranked methods. Its HumanScore is 38.80, compared with 47.25 for HEFT.

\begingroup
\setlength{\tabcolsep}{2.45pt}
\renewcommand{\arraystretch}{1.24}
\centering
\captionof{table}{HumanTracker. Methods without global pose inputs are ranked by overall success. Each motion family reports success, joint-angle MPJPE, and HumanScore.}
\label{tab:full-humantracker}
{\fontsize{7.9}{9.7}\selectfont
\begin{tabular}{@{}rll*{15}{r}@{}}
\toprule
&&&\multicolumn{3}{c}{Overall (2,500)}
  &\multicolumn{3}{c}{Daily (974)}
  &\multicolumn{3}{c}{Highly Dynamic (268)}
  &\multicolumn{3}{c}{Interaction (1,094)}
  &\multicolumn{3}{c}{Ground (164)} \\
\cmidrule(lr){4-6}\cmidrule(lr){7-9}\cmidrule(lr){10-12}\cmidrule(lr){13-15}\cmidrule(lr){16-18}
Rank & Method & Global pose
  & Succ. \(\uparrow\) & MPJPE \(\downarrow\) & HS \(\uparrow\)
  & Succ. & MPJPE & HS
  & Succ. & MPJPE & HS
  & Succ. & MPJPE & HS
  & Succ. & MPJPE & HS \\
\midrule
\multicolumn{18}{@{}l}{\textit{Reference with privileged global pose}} \\
\textemdash & Humanoid-GPT & XY, yaw\textsuperscript{*} & 90.76\%\textsuperscript{\(\dagger\)} & \(\approx\)0.0597 & \(\approx\)54.19 & 94.4\% & 0.046 & 54.7 & 86.9\% & 0.047 & 49.2 & 97.2\% & 0.070 & 56.8 & 32.9\% & 0.216 & 24.9 \\
\midrule
\rowcolor{benchmarklead}
1 & \textbf{HoloMotion-1.5 (ours)} & \textemdash & \textbf{90.36\%} & \textbf{0.0606} & 38.80 & \textbf{95.69\%} & \underline{0.057} & 35.09 & 80.22\% & \textbf{0.060} & 36.06 & 97.26\% & \textbf{0.059} & 45.33 & \textbf{29.27\%} & \textbf{0.148} & 37.29 \\
2 & SONIC 1.1 & \textemdash & \underline{89.56\%\textsuperscript{\(\dagger\)}} & \(\approx\)0.1035 & \(\approx\)44.85 & \underline{94.4\%} & 0.093 & 43.6 & \textbf{84.7\%} & 0.108 & 38.5 & \textbf{97.9\%} & 0.111 & 48.9 & 13.4\% & 0.198 & 33.9 \\
3 & SONIC & \textemdash & 89.40\%\textsuperscript{\(\dagger\)} & \(\approx\)0.1161 & \textbf{\(\approx\)50.07} & 93.8\% & 0.102 & \textbf{49.5} & \underline{82.1\%} & 0.118 & \underline{41.0} & \underline{97.6\%} & 0.128 & \textbf{54.6} & \underline{20.1\%} & 0.231 & 26.5 \\
4 & HEFT & \textemdash & 88.16\% & \underline{0.0621} & 47.25 & 93.2\% & \textbf{0.056} & \underline{46.1} & 75.7\% & \underline{0.061} & 40.5 & 97.5\% & \underline{0.063} & 51.6 & 15.9\% & 0.176 & 31.7 \\
5 & ScaleBFM-M & \textemdash & 86.52\% & 0.0933 & \underline{47.49} & 90.8\% & 0.094 & 44.5 & 72.4\% & 0.091 & 39.3 & 96.5\% & 0.084 & \underline{54.0} & 17.7\% & 0.186 & \underline{43.4} \\
5 & ScaleBFM-XL & \textemdash & 86.52\% & 0.0909 & 46.14 & 90.1\% & 0.093 & 44.2 & 74.3\% & 0.089 & 37.5 & 96.7\% & 0.082 & 51.2 & 17.1\% & \underline{0.158} & 40.8 \\
7 & GRIT v0.0.1 & \textemdash & 85.16\% & 0.1129 & 42.68 & 90.2\% & 0.105 & 40.2 & 65.7\% & 0.115 & \textbf{41.4} & 96.5\% & 0.114 & 48.5 & 11.0\% & 0.241 & 24.9 \\
8 & HoloMotion v1.4.0 & \textemdash & 82.24\% & 0.0922 & 28.28 & 84.0\% & 0.088 & 20.4 & 63.4\% & 0.101 & 27.2 & 94.9\% & 0.091 & 40.4 & 18.3\% & 0.172 & 35.1 \\
9 & TeleopIT & \textemdash & 81.56\% & 0.1044 & 38.61 & 84.0\% & 0.096 & 34.0 & 54.9\% & 0.110 & 35.8 & 96.2\% & 0.105 & 45.7 & 13.4\% & 0.229 & \textbf{47.1} \\
10 & MimicLite-ROA & \textemdash & 80.52\% & 0.0931 & 28.59 & 83.1\% & 0.080 & 24.8 & 59.0\% & 0.112 & 26.9 & 95.2\% & 0.093 & 35.1 & 2.4\% & 0.290 & 26.6 \\
11 & MimicLite-PPO & \textemdash & 80.08\% & 0.0893 & 31.24 & 81.1\% & 0.080 & 28.0 & 58.6\% & 0.096 & 28.1 & 95.2\% & 0.089 & 37.9 & 7.9\% & 0.239 & 20.1 \\
12 & Mimic Lite v1.1 & \textemdash & 78.52\% & 0.0899 & 28.12 & 80.1\% & 0.080 & 25.0 & 51.5\% & 0.114 & 22.8 & 94.4\% & 0.087 & 34.3 & 7.3\% & 0.252 & 23.1 \\
13 & TWIST2 & \textemdash & 67.64\%\textsuperscript{\(\dagger\)} & \(\approx\)0.1148 & \(\approx\)16.71 & 60.1\% & 0.105 & 10.1 & 39.9\% & 0.112 & 16.9 & 91.3\% & 0.111 & 28.3 & 0.0\% & 0.341 & 4.5 \\
14 & GMT & \textemdash & \(\approx\)46.1\%\textsuperscript{\(\dagger\)} & \(\approx\)0.2374 & \(\approx\)5.98 & 17.0\% & 0.250 & 2.4 & 36.2\% & 0.196 & 7.0 & 81.4\% & 0.205 & 11.7 & 0.0\% & 0.456 & 4.0 \\
\bottomrule
\end{tabular}\par}
\smallskip
{\footnotesize\raggedright
\textbf{Metrics.} Overall success divides the total completions across families by 2,500. MPJPE is in radians and HS is HumanScore on a 0--100 scale. \(\dagger\) sums the integer success counts consistent with family rates rounded to 0.1 percentage points. GMT admits 1,153 or 1,154 successes. Where overall MPJPE or HS is unavailable, \(\approx\) marks an estimate weighted by reference frames from rounded family values.

\textbf{Highlights.} Best values are bold and second-best values are underlined. Equal displayed values share highlighting. \textsuperscript{*}Its adapter uses the simulated global root XY position and heading to construct policy inputs~\citep{humantrackerobservations}. Category scores and available exact overall scores are taken from the HumanTracker project page and evaluation repository~\citep{humantrackerwebsite,humantrackerreleased}.\par}
\endgroup

\paragraph{Effect of the specialist policy.}
We evaluate the final policy and its 150K generalist on the same full suites. Specialization adds 13 HumanTracker completions and reduces error from 26.49 to 25.88~mm on LAFAN1 and from 17.86 to 17.66~mm on BONES-SEED.

\paragraph{Evaluation throughput.}
Overlapping staging, scoring, verification, and upload cut evaluation time by 42.6\% on one RTX 5090 (94:44 to 54:22, min:s). The same checkpoint produced matching scores and trace hashes. Storage cache state was uncontrolled.

\vfill\noindent\makebox[\linewidth]{\small\thepage}\par

\clearpage
\thispagestyle{empty}
\subsection{MotionChallenge LAFAN1 results}

LAFAN1 shows whether accurate pose tracking and reliable completion agree on a small set of common references. Across 40 LAFAN1 motions, HoloMotion-1.5 records \textbf{25.88~mm} full-motion XY-aligned error, down 52.8\% from ScaleBFM-M and 82.9\% from HoloMotion v1.4.0. It also leads with early termination (27.17~mm) and with yaw alignment. It completes 28 motions, compared with 33 for MimicLite. Without alignment, MimicLite leads at 627.66~mm and HoloMotion-1.5 records 933.64~mm.

\begingroup
\setlength{\tabcolsep}{2.85pt}
\renewcommand{\arraystretch}{1.13}
\centering
\captionof{table}{MotionChallenge LAFAN1 on the G1 29-DoF track (40 motions). Rank follows the default MPKPE column, \emph{No termination / XY only}.}
\label{tab:full-lafan}
{\fontsize{8.0}{9.0}\selectfont
\begin{tabularx}{\linewidth}{@{}rl l L *{8}{r}@{}}
\toprule
&&&&\multicolumn{3}{c}{No termination (mm)}
  &\multicolumn{3}{c}{Early termination (mm)}&& \\
\cmidrule(lr){5-7}\cmidrule(lr){8-10}
Rank & Model & Type & Organization
  & XY only & XY + yaw & No align.
  & XY only & XY + yaw & No align.
  & Failed & Success \\
\midrule
\rowcolor{benchmarklead}
1 & \textbf{HoloMotion-1.5 (ours)} & MoE & Horizon Robotics & \textbf{25.88} & \textbf{23.32} & \underline{933.64} & \textbf{27.17} & \textbf{23.42} & 760.71 & 12 & 70.0\% (28 / 40) \\
2 & ScaleBFM M & Dense & Shanghai AI Lab \& collaborators & \underline{54.86} & \underline{53.98} & 1692.43 & 46.90 & 46.58 & 1256.21 & 15 & 62.5\% (25 / 40) \\
3 & GRIT 99k & Dense & GRIT Team & 64.17 & 60.80 & 1626.06 & 43.56 & 40.93 & 1405.81 & \underline{8} & \underline{80.0\% (32 / 40)} \\
4 & MimicLite 031b8d67 & Dense & RoboParty & 65.16 & 60.12 & \textbf{627.66} & 34.37 & 32.83 & \underline{488.46} & \textbf{7} & \textbf{82.5\% (33 / 40)} \\
5 & HEFT PMG & Dense & Tsinghua / RobotEra / SQZ & 69.25 & 62.09 & 1240.02 & \underline{33.46} & \underline{28.53} & 963.08 & 10 & 75.0\% (30 / 40) \\
6 & SONIC Original & FSQ & NVIDIA & 90.78 & 85.67 & 2271.21 & 42.31 & 39.89 & 1737.34 & 12 & 70.0\% (28 / 40) \\
7 & SONIC Low-Latency & FSQ & NVIDIA & 118.61 & 110.75 & 1977.69 & 45.14 & 42.66 & 1559.65 & 14 & 65.0\% (26 / 40) \\
8 & SONIC v1.1 & FSQ & NVIDIA & 119.10 & 111.22 & 2165.51 & 44.49 & 40.73 & 1562.93 & 15 & 62.5\% (25 / 40) \\
9 & HoloMotion v1.4.1 & MoE & Horizon Robotics & 141.93 & 119.08 & 1803.97 & 52.64 & 41.88 & 1390.36 & 13 & 67.5\% (27 / 40) \\
10 & TeleopIT G1 & Dense & Westlake / SII & 149.55 & 136.89 & 1792.96 & 53.13 & 51.94 & 1383.16 & 18 & 55.0\% (22 / 40) \\
11 & HoloMotion v1.4.0 & MoE & Horizon Robotics & 151.34 & 124.87 & 1888.29 & 51.63 & 38.06 & 1321.88 & 15 & 62.5\% (25 / 40) \\
12 & HoloMotion v1.3.2 & MoE & Horizon Robotics & 158.35 & 140.54 & 1009.80 & 45.57 & 38.28 & \textbf{335.69} & 17 & 57.5\% (23 / 40) \\
13 & UFO FB-CPR-Aux 192M & FB & RoboParty & 214.92 & 71.00 & 6223.98 & 185.54 & 66.95 & 5066.75 & 15 & 62.5\% (25 / 40) \\
14 & Humanoid-GPT Transformer 216 & Dense & Galbot / Tsinghua \& collaborators & 218.70 & 154.21 & 3806.33 & 105.96 & 51.65 & 2855.36 & 17 & 57.5\% (23 / 40) \\
15 & BFM-Zero & FB & CMU / Meta & 236.07 & 85.32 & 11436.64 & 185.50 & 67.97 & 6436.23 & 22 & 45.0\% (18 / 40) \\
16 & PMT RGMT Transformer 20k & Dense & Mondo Robotics / HKUST \& collaborators & 274.30 & 121.46 & 7753.71 & 196.33 & 43.02 & 7016.29 & 16 & 60.0\% (24 / 40) \\
17 & ProtoMotions G1 BONES & Dense & NVIDIA & 307.68 & 289.70 & 4324.63 & 68.33 & 62.57 & 2659.84 & 28 & 30.0\% (12 / 40) \\
18 & MOSAIC GMT & Dense & BAAI / TUM / Tsinghua \& collaborators & 315.06 & 299.10 & 2531.20 & 62.68 & 61.79 & 1188.50 & 30 & 25.0\% (10 / 40) \\
19 & HoloMotion v1.3.1 & MoE & Horizon Robotics & 348.70 & 260.87 & 2855.48 & 112.36 & 40.42 & 1521.00 & 28 & 30.0\% (12 / 40) \\
20 & TWIST2 1017-20k & Dense & Amazon FAR & 468.67 & 381.67 & 6347.28 & 195.40 & 58.23 & 4627.78 & 34 & 15.0\% (6 / 40) \\
21 & OmniXtreme & Flow & BIGAI / Unitree \& collaborators & 559.16 & 518.94 & 5775.50 & 90.80 & 79.92 & 2186.56 & 40 & 0.0\% (0 / 40) \\
22 & Any2Track Generalist v2 & Dense & Galaxy General Robotics & 610.91 & 553.65 & 7195.16 & 165.70 & 77.40 & 2314.53 & 40 & 0.0\% (0 / 40) \\
\bottomrule
\end{tabularx}\par}
\smallskip
{\footnotesize\raggedright
All error columns report MPKPE. Full-motion errors include post-failure states. Early termination includes the first failure frame. Bold and underlined numbers mark the best and second best, with ties at displayed precision for errors and exact counts for success. The blue row shows our local fixed-order evaluation. The other rows come from MotionChallenge's public results for the same track (revision 86d76f91a6e4)~\citep{motionchallenge}.\par}
\endgroup

\paragraph{Replay repeatability.}
Repeated rollouts tell us whether a score difference can arise with policy weights held fixed. On one RTX 5090, two upstream LAFAN1 evaluations of the same checkpoint completed 28 and 26 of 40 motions. Fixing the floating-point accumulation order in contact and constraint reductions yielded identical per-motion scores and 27 completions in both runs. Our reported runs use this execution mode with Core's scoring rules and independent CPU verification.

\vfill\noindent\makebox[\linewidth]{\small\thepage}\par

\clearpage
\thispagestyle{empty}
\subsection{MotionChallenge BONES-SEED results}

BONES-SEED tests tracking accuracy across a much larger set of references. Across 129,561 BONES-SEED motions, HoloMotion-1.5 records \textbf{17.66~mm} full-motion XY-aligned error, down 35.6\% from HEFT PMG and 59.1\% from HoloMotion v1.4.0. Its early-termination XY-aligned error is 17.55~mm. It completes 128,555 motions. ScaleBFM-M leads on completion with 129,298 motions, and HoloMotion v1.3.2 leads without alignment at 63.49~mm.

\begingroup
\setlength{\tabcolsep}{2.85pt}
\renewcommand{\arraystretch}{1.13}
\centering
\captionof{table}{MotionChallenge BONES-SEED on the G1 29-DoF track (129,561 motions). Rank follows the default MPKPE column, \emph{No termination / XY only}.}
\label{tab:full-seed}
{\fontsize{8.0}{9.0}\selectfont
\begin{tabularx}{\linewidth}{@{}rl l L *{8}{r}@{}}
\toprule
&&&&\multicolumn{3}{c}{No termination (mm)}
  &\multicolumn{3}{c}{Early termination (mm)}&& \\
\cmidrule(lr){5-7}\cmidrule(lr){8-10}
Rank & Model & Type & Organization
  & XY only & XY + yaw & No align.
  & XY only & XY + yaw & No align.
  & Failed & Success \\
\midrule
\rowcolor{benchmarklead}
1 & \textbf{HoloMotion-1.5 (ours)} & MoE & Horizon Robotics & \textbf{17.66} & \textbf{16.81} & \underline{76.64} & \textbf{17.55} & \textbf{16.71} & \underline{75.43} & 1,006 & 99.2\% (128,555 / 129,561) \\
2 & HEFT PMG & Dense & Tsinghua / RobotEra / SQZ & \underline{27.41} & \underline{25.53} & 123.24 & \underline{26.92} & \underline{25.06} & 120.98 & 1,085 & 99.2\% (128,476 / 129,561) \\
3 & MimicLite 031b8d67 & Dense & RoboParty & 30.62 & 29.52 & 109.09 & 29.53 & 28.48 & 105.86 & 1,884 & 98.5\% (127,677 / 129,561) \\
4 & ScaleBFM M & Dense & Shanghai AI Lab \& collaborators & 30.81 & 31.21 & 183.01 & 30.76 & 31.16 & 182.71 & \textbf{263} & \textbf{99.8\% (129,298 / 129,561)} \\
5 & GRIT 99k & Dense & GRIT Team & 32.64 & 32.11 & 164.15 & 32.15 & 31.65 & 162.54 & 619 & 99.5\% (128,942 / 129,561) \\
6 & SONIC Original & FSQ & NVIDIA & 33.48 & 32.51 & 111.65 & 32.93 & 31.96 & 110.50 & 322 & 99.8\% (129,239 / 129,561) \\
7 & SONIC Low-Latency & FSQ & NVIDIA & 34.12 & 32.81 & 191.07 & 33.59 & 32.29 & 189.37 & 371 & 99.7\% (129,190 / 129,561) \\
8 & HoloMotion v1.3.2 & MoE & Horizon Robotics & 35.40 & 31.50 & \textbf{63.49} & 34.36 & 30.50 & \textbf{61.21} & 698 & 99.5\% (128,863 / 129,561) \\
9 & SONIC v1.1 & FSQ & NVIDIA & 35.95 & 32.81 & 160.14 & 35.14 & 32.01 & 157.87 & 675 & 99.5\% (128,886 / 129,561) \\
10 & TeleopIT G1 & Dense & Westlake / SII & 36.52 & 36.03 & 182.97 & 36.19 & 35.72 & 182.02 & \underline{275} & \underline{99.8\% (129,286 / 129,561)} \\
11 & HoloMotion v1.4.0 & MoE & Horizon Robotics & 43.22 & 37.30 & 173.10 & 40.67 & 34.80 & 168.97 & 2,486 & 98.1\% (127,075 / 129,561) \\
12 & HoloMotion v1.4.1 & MoE & Horizon Robotics & 44.77 & 39.88 & 177.43 & 42.20 & 37.38 & 172.90 & 2,460 & 98.1\% (127,101 / 129,561) \\
13 & HoloMotion v1.3.1 & MoE & Horizon Robotics & 46.26 & 39.34 & 110.24 & 41.36 & 34.57 & 99.59 & 3,574 & 97.2\% (125,987 / 129,561) \\
14 & MOSAIC GMT & Dense & BAAI / TUM / Tsinghua \& collaborators & 47.65 & 46.88 & 211.39 & 44.80 & 44.05 & 201.74 & 2,175 & 98.3\% (127,386 / 129,561) \\
15 & Humanoid-GPT Transformer 216 & Dense & Galbot / Tsinghua \& collaborators & 52.89 & 45.75 & 323.26 & 47.35 & 41.26 & 296.20 & 4,333 & 96.7\% (125,228 / 129,561) \\
16 & ProtoMotions G1 BONES & Dense & NVIDIA & 56.83 & 55.13 & 406.81 & 53.79 & 52.22 & 388.03 & 2,677 & 97.9\% (126,884 / 129,561) \\
17 & UFO FB-CPR-Aux 192M & FB & RoboParty & 71.54 & 52.74 & 415.03 & 71.01 & 52.60 & 410.44 & 3,035 & 97.7\% (126,526 / 129,561) \\
18 & PMT RGMT Transformer 20k & Dense & Mondo Robotics / HKUST \& collaborators & 93.07 & 54.84 & 504.24 & 82.20 & 44.33 & 469.19 & 7,075 & 94.5\% (122,486 / 129,561) \\
19 & BFM-Zero & FB & CMU / Meta & 95.30 & 60.42 & 494.44 & 93.52 & 59.75 & 485.41 & 3,287 & 97.5\% (126,274 / 129,561) \\
20 & TWIST2 1017-20k & Dense & Amazon FAR & 101.79 & 56.07 & 551.72 & 93.52 & 48.26 & 511.44 & 6,067 & 95.3\% (123,494 / 129,561) \\
21 & OmniXtreme & Flow & BIGAI / Unitree \& collaborators & 194.21 & 186.15 & 1246.32 & 86.38 & 80.80 & 863.17 & 49,692 & 61.6\% (79,869 / 129,561) \\
22 & Any2Track Generalist v2 & Dense & Galaxy General Robotics & 404.87 & 371.49 & 2112.25 & 143.13 & 113.43 & 954.71 & 101,048 & 22.0\% (28,513 / 129,561) \\
\bottomrule
\end{tabularx}\par}
\smallskip
{\footnotesize\raggedright
All error columns report MPKPE. Full-motion errors include post-failure states. Early termination includes the first failure frame. Bold and underlined numbers mark the best and second best, with ties at displayed precision for errors and exact counts for success. The blue row shows our local fixed-order evaluation. The other rows come from MotionChallenge's public results for the same track (revision 86d76f91a6e4)~\citep{motionchallenge}.\par}
\endgroup
\vfill\noindent\makebox[\linewidth]{\small\thepage}\par
\end{landscape}
\restoregeometry

\begingroup
\setlength{\parskip}{0.26em}
\section{Compute accounting and reproducibility}\label{sec:compute}
\enlargethispage{14pt}

\paragraph{Training budget.}
The runs that produced the evaluated checkpoint had a combined budget of 83 active GPU-hours from prepared references. HoloMotion-1 reports about 9,200 GPU-hours~\citep{holomotion1}. Active time covers collection and updates. Recorded trainer time also includes startup, compilation, and checkpointing (Table~\ref{tab:compute}).

With 32 specialist GPUs running concurrently, the active budgets give a critical path of \(16+2+1+2=21\) hours. The longest specialist logged 2 h 11 min of trainer time. Data preparation, worker setup, scheduling, and evaluation sit outside both clocks.

\noindent\begin{minipage}{\linewidth}
\centering
\captionof{table}{Training budget and recorded GPU-hours for the runs that produced the evaluated checkpoint. Stage~1 counts the full run from which we select the 150K checkpoint.}
\label{tab:compute}
\small
\begin{tabularx}{\linewidth}{Lrr}
\toprule
Training stage & Active budget (GPU-hours) & Trainer GPU-hours \\
\midrule
Stage 1. Shared generalist & 16.00 & 16.81 \\
Stage 2. Residual specialists & 64.00 & 69.38 \\
Stage 3. Global router & 1.00 & 1.26 \\
Stage 3. Residual gate & 2.00 & 2.17 \\
\midrule
Total & \textbf{83.00} & \textbf{89.62} \\
\bottomrule
\end{tabularx}
\end{minipage}

\paragraph{Stage 1 scaling ablation.}
We test whether more GPUs shorten Stage~1 at fixed cost. Runs share a seed, architecture, and 16 active GPU-hours. Each GPU runs 4,096 environments, with 4,096 actor starts and 2,048 dynamics windows per batch. With \(D\) GPUs, global batches scale by \(D\) and base learning rate is \(4\times10^{-4}\sqrt{D}\). At the same GPU-hour budget, the 16-GPU run makes 4.2\% as many actor updates and collects 78.7\% as many transitions as the single-GPU run. The single-GPU run reaches 162,692 steps and leads this ablation. We use its earlier 150K checkpoint to initialize the evaluated specialist bank.

\noindent\begin{minipage}{\linewidth}
\centering
\captionof{table}{Stage 1 scaling ablation at 16 active GPU-hours. Training time is the active budget per GPU. HT is HumanTracker success (\%). LAFAN1 and BONES-SEED use full-motion XY-aligned error (mm).}
\label{tab:stage1-scaling}
\small
\setlength{\tabcolsep}{4pt}
\begin{tabular*}{\linewidth}{@{\extracolsep{\fill}}rrrrrrr@{}}
\toprule
GPUs & Training (h) & Transitions & Actor updates & HT & LAFAN1 & BONES-SEED \\
\midrule
1 & 16 & 666.4M & 64,551 & 89.52 & 26.52 & 17.96 \\
2 & 8 & 648.7M & 31,163 & 88.60 & 28.89 & 19.43 \\
4 & 4 & 631.5M & 14,912 & 88.32 & 29.77 & 21.27 \\
8 & 2 & 538.3M & 6,068 & 86.32 & 33.18 & 22.74 \\
16 & 1 & 524.5M & 2,698 & 82.88 & 42.10 & 26.24 \\
\bottomrule
\end{tabular*}
\end{minipage}

\paragraph{Expert computation.}
We measure expert execution across the training schedule, from all MLPs in dense routing to two per token at hard Top-2. Separate tests measure adaptive dispatch, which groups nonzero pairs at mean support up to 12 and per-token support up to 32. On one RTX 5090 with 4,096 synthetic BF16 tokens, frozen-block forward and backward run \(8.7\times\), \(8.6\times\), \(8.2\times\), and \(7.5\times\) faster at mean supports of 2.7, 3.7, 6.6, and 8.8. Dense fallback adds 2.5--2.6\% to block time. With trainable experts and 24-step rollouts, sparse dispatch and batched updates cut actor-and-critic time from 6.14 to 1.27~s at mean support 3.2. At full support, optimizer batching cuts update time from 6.07 to 4.51~s.

\section{Discussion and Future Work}

The directions below extend the training strategy and framework introduced here. We reserve the broader HoloMotion roadmap for separate discussion.

\paragraph{Differentiating through contact.}
The learned dynamics model makes imagined updates inexpensive. Around impacts and changes in support, its gradients can differ from those of the simulator. \(\nabla\)Mimic differentiates through the contact solver and controls extreme gradients with per-sample clipping and regression to gradient-derived action targets~\citep{nablamimic}. One direction is to test simulator gradients on contact-rich specialist groups while keeping the observation, reward, and controller fixed. The comparison would show whether better gradients around contact improve tracking enough to offset their extra computation.

\paragraph{Pipelining training across GPUs.}
Section~\ref{sec:compute} tests synchronous data parallelism. XMimic offers a different way to use multiple devices: one GPU collects experience, another updates the world model and critic, a third updates the policy, and a fourth runs evaluations that guide resets~\citep{xmimic}. Overlap can shorten an iteration, but it introduces delay between data collection and policy or world-model updates. For our Transformer, the question is whether a pipeline preserves tracking quality as it shortens elapsed training time. A comparison should hold the batch sizes and update ratios fixed and report time to the same tracking score along with total GPU-hours.

\paragraph{Allocating practice to difficult motions.}
Large catalogs make it hard to check every clip for failures before choosing what to practice. BlinkMimic and BlinkMimic~2 motivate using earlier rollouts to revisit difficult motion segments~\citep{blinkmimic,blinkmimic2}. The open question in a rotating reference bank is how to give those segments more practice while still bringing new clips into the working set. We would measure performance on difficult motions together with distinct-frame coverage and held-out success.

\section{Conclusion}

HoloMotion-1.5 completes 90.36\% of held-out HumanTracker motions, the highest success rate among compared policies without global pose inputs. On MotionChallenge, it achieves the lowest full-motion XY-aligned error on both tracks, with 25.88~mm on LAFAN1 and 17.66~mm on BONES-SEED. These errors are 53\% and 36\% below the previous published bests, respectively.

The three-stage framework scales both the motion catalog and the expert bank of a humanoid tracker. It learns shared feedback once, trains residual specialists in small parallel groups to limit expert computation, and fits a reference-only global router and gate over the frozen bank. Compact streaming bounds reference memory, and deployment uses two of 1,024 residual experts per step. The 0.4B tracker trains on a catalog of over 10,000 hours with a budget of 83 active GPU-hours.

\endgroup

\bibliographystyle{unsrtnat}
\bibliography{references}

@misc{holomotion1,
  author = {Maiyue Chen and Kaihui Wang and Bo Zhang and Xihan Ma and Zhiyuan Yang and Yi Ren and Qijun Huang and Zihao Zhu and Yucheng Wang and Zhizhong Su},
  title = {{HoloMotion-1} Technical Report},
  year = {2026},
  howpublished = {\href{https://arxiv.org/abs/2605.15336}{arXiv:2605.15336}}
}

@misc{supermimic,
  author = {Qiayuan Liao and Yiyang Shao and Junfeng Long and Kevin Zakka and Roberto Horowitz and Pieter Abbeel and Koushil Sreenath},
  title = {{SuperMimic}: Scaling Humanoid Whole-Body Control with One Differentiable World Model},
  year = {2026},
  howpublished = {\href{https://arxiv.org/abs/2610.28480}{arXiv:2610.28480}},
  note = {\href{https://github.com/HybridRobotics/supermimic/blob/5efbf114e11f725c31d83e217be987c13c0e4ce2/src/supermimic/representations.py\#L95-L100}{Official implementation}}
}

@misc{tfdata,
  author = {Derek G. Murray and Jiri Simsa and Ana Klimovic and Ihor Indyk},
  title = {{tf.data}: A Machine Learning Data Processing Framework},
  year = {2021},
  url = {https://arxiv.org/abs/2101.12127}
}

@inproceedings{ffcv,
  author = {Guillaume Leclerc and Andrew Ilyas and Logan Engstrom and Sung Min Park and Hadi Salman and Aleksander Madry},
  title = {{FFCV}: Accelerating Training by Removing Data Bottlenecks},
  booktitle = {CVPR},
  year = {2023},
  note = {\href{https://arxiv.org/abs/2306.12517}{arXiv:2306.12517}}
}

@misc{mjlabloader,
  author = {{mjlab Contributors}},
  title = {{mjlab\_research} Motion Reference Loader},
  year = {2026},
  howpublished = {\href{https://github.com/HybridRobotics/mjlab_research/blob/dd51a9fcc98791a2f0b018b3c7f71a3eefd5ecdd/src/mjlab/tasks/tracking/mdp/commands.py\#L47-L121}{Source revision dd51a9f}}
}

@misc{upcycling,
  author = {Aran Komatsuzaki and Joan Puigcerver and James Lee-Thorp and Carlos Riquelme Ruiz and Basil Mustafa and Joshua Ainslie and Yi Tay and Mostafa Dehghani and Neil Houlsby},
  title = {Sparse Upcycling: Training Mixture-of-Experts from Dense Checkpoints},
  year = {2023},
  url = {https://arxiv.org/abs/2212.05055}
}

@misc{deepseekmoe,
  author = {Damai Dai and Chengqi Deng and Chenggang Zhao and R. X. Xu and Huazuo Gao and others},
  title = {{DeepSeekMoE}: Towards Ultimate Expert Specialization in Mixture-of-Experts Language Models},
  year = {2024},
  howpublished = {\href{https://arxiv.org/abs/2401.06066}{arXiv:2401.06066}}
}

@article{switchtransformer,
  author = {William Fedus and Barret Zoph and Noam Shazeer},
  title = {Switch Transformers: Scaling to Trillion Parameter Models with Simple and Efficient Sparsity},
  journal = {Journal of Machine Learning Research},
  year = {2022},
  volume = {23},
  number = {120},
  pages = {1--39},
  url = {https://www.jmlr.org/papers/v23/21-0998.html}
}

@misc{bumblebee,
  author = {Yuxuan Wang and Ming Yang and Ziluo Ding and Yu Zhang and Weishuai Zeng and Xinrun Xu and Haobin Jiang and Zongqing Lu},
  title = {From Experts to a Generalist: Toward General Whole-Body Control for Humanoid Robots},
  year = {2025},
  howpublished = {\href{https://arxiv.org/abs/2506.12779}{arXiv:2506.12779}}
}

@misc{btx,
  author = {Sainbayar Sukhbaatar and Olga Golovneva and Vasu Sharma and Hu Xu and Xi Victoria Lin and Baptiste Rozi{\`e}re and Jacob Kahn and Daniel Li and Wen-tau Yih and Jason Weston and Xian Li},
  title = {{Branch-Train-MiX}: Mixing Expert {LLMs} into a Mixture-of-Experts {LLM}},
  year = {2024},
  howpublished = {\href{https://arxiv.org/abs/2403.07816}{arXiv:2403.07816}}
}

@inproceedings{residualrl,
  author = {Tobias Johannink and Shikhar Bahl and Ashvin Nair and Jianlan Luo and Avinash Kumar and Matthias Loskyll and Juan Aparicio Ojea and Eugen Solowjow and Sergey Levine},
  title = {Residual Reinforcement Learning for Robot Control},
  booktitle = {ICRA},
  year = {2019},
  url = {https://arxiv.org/abs/1812.03201}
}

@misc{xmimic,
  author = {Nico Bohlinger and Jan Peters},
  title = {{XMimic} and {XSoma}: Learning Humanoid Motion Tracking and Locomotion in Under a Minute},
  year = {2026},
  howpublished = {\href{https://arxiv.org/abs/2611.20452}{arXiv:2611.20452}}
}

@misc{instanttrack,
  author = {Huy Quang Dao},
  title = {{InstantTrack} Whole-Body Motion Tracking Implementation},
  year = {2026},
  url = {https://github.com/Huy-Quang-Dao/instantTrack}
}

@misc{blinkmimic,
  author = {Fuyi Chen},
  title = {{BlinkMimic}: Training a Humanoid Motion Tracker in Minutes on a Single GPU},
  year = {2026},
  howpublished = {\href{https://arxiv.org/abs/2611.16422}{arXiv:2611.16422}}
}

@misc{blinkmimic2,
  author = {{BlinkMimic 2 Contributors}},
  title = {{BlinkMimic 2} Whole-Body Motion Tracking Implementation},
  year = {2026},
  url = {https://github.com/jyl681/blinkmimic2}
}

@misc{nablamimic,
  author = {Weyl Lu and Yubei Chen},
  title = {{$\nabla$Mimic}: Ultra-Efficient Humanoid Control with Differentiable Simulation},
  year = {2026},
  howpublished = {\href{https://arxiv.org/abs/2610.41654}{arXiv:2610.41654}}
}

@inproceedings{entmax,
  author = {Ben Peters and Vlad Niculae and Andr{\'e} F. T. Martins},
  title = {Sparse Sequence-to-Sequence Models},
  booktitle = {Proceedings of the 57th Annual Meeting of the Association for Computational Linguistics},
  year = {2019},
  url = {https://aclanthology.org/P19-1146/}
}

@misc{humantracker,
  author = {Dairu Liu and Zekun Qi and Jiayu Zeng and Ruixi Yu and Yu Guan and others},
  title = {{HumanTracker}: Towards Comprehensive and Human-Aligned Motion Tracking Benchmark},
  year = {2026},
  howpublished = {\href{https://arxiv.org/abs/2608.13555}{arXiv:2608.13555}}
}

@misc{motionchallenge,
  author = {{MotionChallenge Project}},
  title = {{MotionChallenge} Leaderboard and Evaluation Protocol},
  year = {2026},
  url = {https://motionchallenge-project.github.io/motionchallenge/}
}

@misc{motionchallengecore,
  author = {{MotionChallenge Project}},
  title = {{MotionChallenge-Core}: An Open Benchmark for Humanoid Motion Tracking},
  year = {2026},
  url = {https://github.com/MotionChallenge-Project/MotionChallenge-Core}
}

@misc{motionchallengeprotocol,
  author = {{MotionChallenge Project}},
  title = {{MotionChallenge} Evaluation Standards},
  year = {2026},
  url = {https://motionchallenge-project.github.io/motionchallenge/protocol/}
}

@misc{humantrackerdata,
  author = {{HumanTracker Contributors}},
  title = {{HumanTracker} Evaluation Test Split},
  year = {2026},
  url = {https://huggingface.co/datasets/GalaxyGeneralRobotics/HumanTracker}
}

@misc{humantrackerobservations,
  author = {{HumanTracker Contributors}},
  title = {{HumanTracker} Native {Humanoid-GPT} Runtime},
  year = {2026},
  url = {https://github.com/GalaxyGeneralRobotics/HumanTracker/blob/main/src/humantracker/eval/native/hgpt/runtime.py}
}

@misc{humantrackerreleased,
  author = {{HumanTracker Contributors}},
  title = {{HumanTracker} Released-Policy Evaluation Results},
  year = {2026},
  url = {https://github.com/GalaxyGeneralRobotics/HumanTracker/blob/main/src/humantracker/eval/README.md#released-policy-results}
}

@misc{humantrackerwebsite,
  author = {{HumanTracker Contributors}},
  title = {{HumanTracker} Zero-Shot Tracking Results},
  year = {2026},
  url = {https://dairuliu.github.io/humantracker/#results}
}

@article{lafan1,
  author = {F{\'e}lix G. Harvey and Mike Yurick and Derek Nowrouzezahrai and Christopher Pal},
  title = {Robust Motion In-betweening},
  journal = {ACM Transactions on Graphics},
  volume = {39},
  number = {4},
  year = {2020},
  doi = {10.1145/3386569.3392480},
  url = {https://github.com/ubisoft/ubisoft-laforge-animation-dataset}
}

@inproceedings{amass,
  author = {Naureen Mahmood and Nima Ghorbani and Nikolaus F. Troje and Gerard Pons-Moll and Michael J. Black},
  title = {{AMASS}: Archive of Motion Capture as Surface Shapes},
  booktitle = {International Conference on Computer Vision},
  year = {2019},
  url = {https://amass.is.tue.mpg.de/}
}

@misc{bonesseed,
  author = {{Bones Studio}},
  title = {{BONES-SEED}: Skeletal Everyday Embodiment Dataset},
  year = {2026},
  url = {https://huggingface.co/datasets/bones-studio/seed}
}

@misc{chingmudata,
  author = {{ChingMu}},
  title = {{MotionDecode}: ChingMu 1000-Hour Embodied Motion Dataset},
  year = {2026},
  url = {https://chingmudata.github.io/MotionDecode/}
}

@misc{hiphi,
  author = {Jiahao Ji and Ji Ma and Runhan Zhang and Runyi Yu and Wenjia Wang and Weiheng Chi and Qianqian Peng and Weichao Yan and Yongfei Gu and Ye Tian and Ting Wu and Longwei Li and Chun Yuan and Ruoli Dai and Lei Han},
  title = {{HiPHI}: A Large-Scale Benchmark for High-Precision Human Motion and Object-Interaction},
  year = {2026},
  howpublished = {\href{https://arxiv.org/abs/2608.16222}{arXiv:2608.16222}}
}

@inproceedings{motionmillion,
  author = {Ke Fan and Shunlin Lu and Minyue Dai and Runyi Yu and Lixing Xiao and Zhiyang Dou and Junting Dong and Lizhuang Ma and Jingbo Wang},
  title = {Go to Zero: Towards Zero-shot Motion Generation with Million-scale Data},
  booktitle = {International Conference on Computer Vision},
  year = {2025},
  url = {https://vankouf.github.io/MotionMillion/}
}

@misc{omnihumo,
  author = {Yiheng Li and Zhuo Li and Ruibing Hou and Yingjie Chen and Hong Chang and Hao Liu and Shiguang Shan},
  title = {{AnyMo}: Scaling Any-Modality Conditional Motion Generation with Masked Modeling},
  year = {2026},
  howpublished = {\href{https://arxiv.org/abs/2605.29488}{arXiv:2605.29488}}
}

@misc{sonic,
  author = {Zhengyi Luo and Ye Yuan and Tingwu Wang and Chenran Li and others},
  title = {{SONIC}: Supersizing Motion Tracking for Natural Humanoid Whole-Body Control},
  year = {2025},
  howpublished = {\href{https://arxiv.org/abs/2511.07820}{arXiv:2511.07820}}
}

@misc{movenszero,
  author = {Wendong Bu and Tinghong Ye and Qizhou Wang and Jiacheng Li and others},
  title = {{Movens-Zero}: Scaling Humanoid Motion Foundation Models with Large-Scale Human Videos},
  year = {2026},
  howpublished = {\href{https://arxiv.org/abs/2610.41877}{arXiv:2610.41877}}
}

@misc{scalebfm,
  author = {Weishuai Zeng and Kangning Yin and Xiaojie Niu and Shunlin Lu and others},
  title = {Scaling Behavior Foundation Model for Humanoid Robots},
  year = {2026},
  howpublished = {\href{https://arxiv.org/abs/2607.15163}{arXiv:2607.15163}}
}

@misc{gmr,
  author = {Jo{\~a}o Pedro Ara{\'u}jo and Yanjie Ze and Pei Xu and Jiajun Wu and C. Karen Liu},
  title = {Retargeting Matters: General Motion Retargeting for Humanoid Motion Tracking},
  year = {2025},
  howpublished = {\href{https://arxiv.org/abs/2510.02252}{arXiv:2510.02252}}
}

@misc{somaretgt,
  author = {{NVIDIA}},
  title = {{SOMA Retargeter}},
  year = {2026},
  howpublished = {\href{https://github.com/NVIDIA/soma-retargeter}{Project repository}}
}

@misc{lafanretarget,
  author = {{Unitree Robotics}},
  title = {{LAFAN1} Retargeting Dataset},
  year = {2025},
  howpublished = {\href{https://huggingface.co/datasets/lvhaidong/LAFAN1_Retargeting_Dataset/tree/ce1572906efe6157840e8474d5a0d7aa87481e74}{Dataset release ce15729}}
}

@article{smpl,
  author = {Matthew Loper and Naureen Mahmood and Javier Romero and Gerard Pons-Moll and Michael J. Black},
  title = {{SMPL}: A Skinned Multi-Person Linear Model},
  journal = {ACM Transactions on Graphics},
  volume = {34},
  number = {6},
  year = {2015},
  url = {https://virtualhumans.mpi-inf.mpg.de/papers/SMPL15/SMPL15.pdf}
}

@misc{holosmpl,
  author = {{Horizon Robotics}},
  title = {{HoloSMPL}: A Common Format for Human Motion Clips},
  year = {2026},
  url = {https://github.com/HorizonRobotics/HoloMotion/tree/master/holosmpl}
}

@misc{holoretarget,
  author = {{Horizon Robotics}},
  title = {{HoloRetarget}: Motion Retargeting for Training and Live Teleoperation},
  year = {2026},
  howpublished = {\href{https://github.com/HorizonRobotics/HoloMotion/tree/master/holoretarget}{Project documentation}}
}

@misc{spdl,
  author = {Moto Hira and Christian Puhrsch and Valentin Andrei and Roman Malinovskyy and Gael Le Lan and Abhinandan Krishnan and Joseph Cummings and Victor Bourgin and Olga Gerasimova and Miguel Martin and Gokul Gunasekaran and Yuta Inoue and Alex J. Turner and Raghuraman Krishnamoorthi},
  title = {Scalable and Performant Data Loading},
  year = {2025},
  howpublished = {\url{https://arxiv.org/abs/2504.20067}}
}

@misc{lerobotv3,
  author = {Francesco Capuano and Michel Aractingi and Quentin Lhoest and Caroline Pascal and Pepijn Kooijmans and Jade Choghari and Remi Cadene and Simon Aliberts and Adil Zouitine and Martino Russi and Steven Palma},
  title = {{LeRobotDataset v3.0}: Bringing Large-Scale Datasets to {LeRobot}},
  year = {2025},
  howpublished = {\url{https://huggingface.co/blog/lerobot-datasets-v3}}
}

@misc{reverb,
  author = {Albin Cassirer and Gabriel Barth-Maron and Eugene Brevdo and Sabela Ramos and Toby Boyd and Thibault Sottiaux and Manuel Kroiss},
  title = {Reverb: A Framework for Experience Replay},
  year = {2021},
  howpublished = {\url{https://arxiv.org/abs/2102.04736}}
}

@inproceedings{flexolmo,
  author = {Weijia Shi and Akshita Bhagia and Kevin Farhat and Niklas Muennighoff and Jacob Morrison and Pete Walsh and Dustin Schwenk and Shayne Longpre and Jake Poznanski and Allyson Ettinger and Daogao Liu and Margaret Li and Dirk Groeneveld and Mike Lewis and Wen-tau Yih and Luca Soldaini and Kyle Lo and Noah A. Smith and Luke Zettlemoyer and Pang Wei Koh and Hannaneh Hajishirzi and Ali Farhadi and Sewon Min},
  title = {{FlexOlmo}: Open Language Models for Flexible Data Use},
  booktitle = {Advances in Neural Information Processing Systems},
  year = {2025},
  url = {https://arxiv.org/abs/2507.07024}
}

@inproceedings{bts,
  author = {Qizhen Zhang and Prajjwal Bhargava and Chloe Bi and Chris X. Cai and Jakob Nicolaus Foerster and Jeremy Fu and Punit Singh Koura and Ruan Silva and Sheng Shen and Emily Dinan and Suchin Gururangan and Mike Lewis},
  title = {{BTS}: Harmonizing Specialized Experts into a Generalist {LLM}},
  booktitle = {Proceedings of the 2025 Conference on Empirical Methods in Natural Language Processing},
  year = {2025},
  url = {https://aclanthology.org/2025.emnlp-main.347/}
}

@article{heald2021,
  author = {James B. Heald and M{\'a}t{\'e} Lengyel and Daniel M. Wolpert},
  title = {Contextual inference underlies the learning of sensorimotor repertoires},
  journal = {Nature},
  year = {2021},
  volume = {600},
  pages = {489--493},
  url = {https://doi.org/10.1038/s41586-021-04129-3}
}

@inproceedings{correia2020sparsemarginalization,
  title = {Efficient Marginalization of Discrete and Structured Latent Variables via Sparsity},
  author = {Correia, Gon{\c{c}}alo M. and Niculae, Vlad and Aziz, Wilker and Martins, Andr{\'e} F. T.},
  booktitle = {Advances in Neural Information Processing Systems},
  year = {2020},
  url = {https://proceedings.neurips.cc/paper/2020/file/887caadc3642e304ede659b734f79b00-Paper.pdf}
}

@misc{heft,
  title = {{HEFT}: Heavy-Payload Full-size Humanoid Teleoperation with Privileged Motion Guidance and Windowed Payload Curriculum},
  author = {Liu, Chenxin and Lu, Qingzhou and Yang, Guangxiao and Shi, Xuanyang and Yang, Chenghan and Guo, Yanjiang and Chen, Jianyu},
  year = {2026},
  howpublished = {arXiv:2607.02332},
  url = {https://arxiv.org/abs/2607.02332},
  note = {G1 observation configuration: \url{https://github.com/Axellwppr/motion_tracking/blob/main/cfg/task/common/tracking_base.yaml}}
}
\end{document}